\def\year{2021}\relax
\documentclass[letterpaper]{article} % DO NOT CHANGE THIS
\usepackage{aaai21}  % DO NOT CHANGE THIS
\usepackage{times}  % DO NOT CHANGE THIS
\usepackage{helvet} % DO NOT CHANGE THIS
\usepackage{courier}  % DO NOT CHANGE THIS
\usepackage[hyphens]{url}  % DO NOT CHANGE THIS
\usepackage{graphicx} % DO NOT CHANGE THIS
\usepackage{natbib}  % DO NOT CHANGE THIS AND DO NOT ADD ANY OPTIONS TO IT
\usepackage{caption} % DO NOT CHANGE THIS AND DO NOT ADD ANY OPTIONS TO IT
\usepackage{amsthm}
\usepackage{amsmath}
\usepackage{bm}
\theoremstyle{definition}
\newtheorem{definition}{Definition}
\newtheorem{theorem}{Theorem}
\usepackage{amsfonts}
\usepackage{algorithm}
\usepackage{algpseudocode}
\usepackage{tabularx}
\usepackage{makecell}
\usepackage{booktabs, threeparttable}
\usepackage{microtype}
\usepackage{graphicx}
\usepackage{subcaption}
\usepackage{multirow}
\usepackage[usenames,dvipsnames,svgnames,table]{xcolor}
\usepackage{colortbl}
\usepackage{hhline}
\usepackage[symbol]{footmisc}
\usepackage{cleveref}
\usepackage[switch]{lineno}

\newcommand{\fw}{{ESCAPED}}
\newcommand{\pfw}{{UAP}}
\newcommand{\norm}[1]{\left\lVert#1\right\rVert}
\newcolumntype{C}{>{\centering\columncolor{gray!5!white}\arraybackslash}X}
\newcolumntype{L}{>{\raggedright\arraybackslash}X}
\newcolumntype{R}{>{\centering\columncolor{gray!30!white}\arraybackslash}p{0.2\textwidth}}
\newcolumntype{G}{>{\centering\columncolor{gray!10!white}\arraybackslash}X}
\newcolumntype{F}{>{\centering\columncolor{gray!30!white}\arraybackslash}X}

\newcommand{\midsepremove}{\aboverulesep = 0mm \belowrulesep = 0mm} 

\newlength\myindent
\crefformat{footnote}{#2\footnotemark[#1]#3}
\title{ESCAPED: Efficient Secure and Private Dot Product Framework for Kernel-based Machine Learning Algorithms with Applications in Healthcare}
\author{
    Ali Burak \"Unal \textsuperscript{\rm 1}, Mete Akg\"un \textsuperscript{\rm 1,2}, Nico Pfeifer \textsuperscript{\rm 1,3}
}
\affiliations{
    \textsuperscript{\rm 1}Methods in Medical Informatics, Department of Computer Science, University of Tuebingen, Germany\\
    \textsuperscript{\rm 2}Translational Bioinformatics, University Hospital Tuebingen, Tuebingen, Germany\\
    \textsuperscript{\rm 3}Statistical Learning in Computational Biology, Max Planck Institute for Informatics, Saarbr\"ucken, Germany\\
    
    \{uenal,akguen,pfeifer\}@informatik.uni-tuebingen.de

}

\begin{document}

\maketitle

\begin{abstract}
To train sophisticated machine learning models one usually needs many training samples. Especially in healthcare settings these samples can be very expensive, meaning that one institution alone usually does not have enough on its own. Merging privacy-sensitive data from different sources is usually restricted by data security and data protection measures. This can lead to approaches that reduce data quality by putting noise onto the variables (e.g., in $\epsilon$ differential privacy) or omitting certain values (e.g., for $k$-anonymity). Other measures based on cryptographic methods can lead to very time-consuming computations, which is especially problematic for larger multi-omics data. We address this problem by introducing ESCAPED, which stands for Efficient SeCure And PrivatE Dot product framework, enabling the computation of the dot product of vectors from multiple sources on a third-party, which later trains kernel-based machine learning algorithms, while neither sacrificing privacy nor adding noise. We evaluated our framework on drug resistance prediction for HIV-infected people and multi-omics dimensionality reduction and clustering problems in precision medicine. In terms of execution time, our framework significantly outperforms the best-fitting existing approaches without sacrificing the performance of the algorithm. Even though we only show the benefit for kernel-based algorithms, our framework can open up new research opportunities for further machine learning models that require the dot product of vectors from multiple sources.
\end{abstract}

\section{Introduction}
In the era of data, the same kind of data is produced by multiple sources. Utilizing such variety of sources is one of the easiest ways to satisfy the eagerness of machine learning algorithms to data. Often, one can train a machine learning model on the pooled data from different sources to get high accuracy on a particular prediction task. However, gathering data can compromise the sensitive information of the samples in the data. \citet{ayday2015whole} showed that genomic data can be used to infer the physical and mental health condition of a patient with the help of the information of patient's lifestyle and environment. Furthermore, \citet{kale2017utility} introduced a method to keep kinship private in an anonymously released genomic dataset, from which such information could otherwise be inferred. Several studies \citep{lunshof2008genetic,azencott2018machine,bonomi2020privacy} discussed various privacy issues from different aspects occurring in studies using medical data.

%In the era of data, there is an enormous number of data generation by many different areas. One of those areas is the health. Multiple sources, such as hospitals and cancer centers, produce the same type of data locally. These sources analyze their own patients and store the data of them for a cancer type. Utilizing those data sources, machine learning algorithms achieve great success. However, employing data from multiple sources raises considerable privacy concerns. Releasing the data in plaintext domain compromises the private information of the data. On the other hand, employing different privacy sacrifices the utility of the data for the sake of privace. In order not to affect the performance of the algorithms while preserving the privacy of the data, heavy encryption techniques, like homomorphic encryption, can be applied but they significantly affect the performance of the algorithm in terms of execution time. In this paper, we propose a framework to enable the computation of the dot product of vectors belonging to different source in a third-party without sacrificing the privacy of the data and decreasing the performance of the algorithm in terms of both the execution time and the accuracy. We utilize the framework to train support vector machines (SVM) model.

Throughout this paper, we refer to a source having data or an entity performing computation as ``{\it party}''. One class of machine learning methods that usually require gathering the whole data, is kernel-based learning methods. To train such a model privately, one of the architectural models in the literature is the distributed model where each party in the computation has its own data and the desired kernel matrix contains the whole data that all parties have.
\citet{vaidya2008privacy} proposed an algorithm utilizing such a model to compute the gram matrix of the whole data belonging to the parties in the computation and train a support vector machine (SVM) privately afterwards. The disadvantage of the proposed algorithm is that it focuses only on binary vectors since it utilizes private set intersection to compute the dot product. 
%Moreover, the distributed model does not fit into our scenario where we have a single party computing the algorithm. 
In addition to the distributed model, there is also the outsourced model where the data is outsourced after encryption and then these encrypted data are used to train a kernel-based machine learning method. \citet{liu2015encrypted} proposed an approach to use an SVM on the encrypted outsourced data. Due to the nature of encryption, the proposed approach is very time consuming. \citet{zhang2017secure} introduced a key-switching \cite{zhou2014efficient} based secure dot product calculation method. The basic idea is to change the key of the dot product of the vectors, which is originally the combination of the keys utilized to encrypt these vectors, to the key of the server. \citet{unaletal2019} demonstrated the inefficiency of this method and proposed a randomized encoding based framework to compute the dot product of the vectors of two parties in a third-party to train an SVM model. However, the framework is not extendable to more than two data sources since this would compromise the data due to the nature of elementwise multiplication of the vectors, which they use to compute the dot product. Furthermore, due to the same reason, their approach has a potential privacy leakage for binary encoded data even for the case with two data sources. We will show that our approach outperforms their framework in such a scenario.
Moreover, the randomized encoding itself \cite{applebaum2006computationally} is independently applicable to our scenario. The authors claimed that any function expressed by a logarithmic depth arithmetic circuit can be encoded by randomized encoding. In this work, we implemented and applied the randomized encoding based approach and show that it is not as efficient as our framework in terms of the communication cost.

%In this paper, we address the privacy problem of data gathering. We come up with a framework, namely extended secure dot product (\fw), by extending the approach of \citet{unaletal2019} and improving their encoding scheme to allow more than two input-parties. \fw{} enables the efficient computation of a dot product of vectors from multiple sources, which we call {\it input-parties}, without neither gathering the data in plaintext domain nor compromising the privacy of the data. It allows a third-party, which we call {\it function-party}, to privately obtain the gram matrix of input-parties' vectors of size larger than $1$. Then the function-party trains a kernel-based machine learning method. We utilized \fw{} to train an SVM in the supervised learning experiments and perform privacy preserving multi-omics dimensionality reduction/clustering in the unsupervised learning experiments.

In this paper, we address the privacy problem of data gathering for dot product based algorithms such as kernel-based learning methods. We first implement and apply one of the fastest encoding in the literature, namely the randomized encoding, to our scenario. Due to the inefficiency of the randomized encoding based approach, we come up with a new encoding scheme enabling the secure and private computation of the dot product of vectors. On top of it, we build a new framework, called efficient secure and private dot product (\fw{}), which allows multiple data sources, which we call {\it input-parties}, to involve in the computation of the dot product. \fw{} allows a third-party, which we call {\it function-party}, to privately obtain the dot product of input-parties' vectors of size larger than $1$ while neither gathering the data in plaintext domain nor compromising the privacy of the data. Then, the function-party trains a kernel-based machine learning method. We utilized \fw{} to predict personalized treatment recommendation for HIV-infected patients in the supervised learning experiment and to perform privacy preserving multi-omics dimensionality reduction and clustering in the unsupervised learning experiments. To the best of our knowledge, this is the first study enabling the privacy preserving multi-omics dimensionality reduction and clustering.

%In this work, we extend that approach and improve the encoding scheme to enable more than two parties with data to participate in the computation of the dot product without sacrificing privacy. Moreover, the randomized encoding itself \cite{applebaum2006computationally} is independently applicable to our scenario. The authors claimed that any function expressed by a logarithmic depth arithmetic circuit can be encoded by the randomized encoding. In this work, we implemented and applied the randomized encoding based approach. We show that it is not as efficient as our framework in terms of the communication cost.

%In this paper, we address the privacy problem of data gathering. We propose a framework, namely extended secure dot product (\fw), to enable the computation of a dot product of vectors from different sources, which we call {\it input-parties}, without neither gathering the data in plaintext domain nor compromising the privacy of the data. \fw{} allows a third-party, which we call {\it function-party}, to privately obtain the gram matrix of input-parties' vectors of size larger than $1$. Then the function-party trains a kernel-based machine learning method. We utilized \fw{} to train an SVM in the supervised learning experiments and perform privacy preserving multi-omics dimensionality reduction/clustering in the unsupervised learning experiments.

%%---------------------------- Background ----------------------------------------------
\section{Background} \label{sec:background}

\subsection{Radial Basis Function Kernel}
Among the kernel functions, the radial basis function (RBF) kernel is one of the most effective and widely used kernels \cite{scholkopf2002learning,kauppi2015towards,pfeifer2008multiple,zhang2004tumor}. The computation of the RBF kernel for samples $x,y \in \mathbb{R}^n$ can be expressed based on only the dot product of these samples. The formula is as follows:
\begin{equation} \label{eq:rbfdotproduct}
K(x,y) = \exp\Bigg(-\dfrac{\norm{\langle x,x\rangle - 2 \langle x,y\rangle + \langle y,y\rangle}^2}{2 \sigma^2}\Bigg)
\end{equation}
where ``$\langle \cdot,\cdot\rangle$'' represents the dot product of vectors and $\sigma$ is the parameter that adjusts the similarity level between the samples. Equation \ref{eq:rbfdotproduct} indicates that the gram matrix is enough to compute the RBF kernel. We benefit from such computation to obtain the RBF kernel matrix in \fw.

%Among the kernel functions, the radial basis function (RBF) kernel is one of the most popular kernels \cite{scholkopf2002learning}. Its effectiveness makes it applicable to large number of areas \cite{kauppi2015towards,pfeifer2008multiple,zhang2004tumor}. The computation of the RBF kernel for samples $x,y \in \mathbb{R}^n$ can be expressed based on only the dot product of these samples. The formula is as follows:
%\begin{equation} \label{eq:rbfdotproduct}
%K(x,y) = \exp\Bigg(-\dfrac{\norm{\langle x,x\rangle - 2 \langle x,y\rangle + \langle y,y\rangle}^2}{2 \sigma^2}\Bigg)
%\end{equation}
%where ``$\langle \cdot,\cdot\rangle$'' represents the dot product of vectors and $\sigma$ is the parameter that adjusts the similarity level between the samples. Equation \ref{eq:rbfdotproduct} indicates that the gram matrix is enough to compute the RBF kernel. We benefit from such computation to obtain the RBF kernel matrix in \fw.

%---------------------------------------------------------------------------------------------------------------

\subsection{Randomized Encoding}

Randomized encoding (RE) is designed to hide the input value $s$ in the computation of a function $f(s)$ by encoding the function with a randomized function $\hat{f}(s;r)$, where $r$ is a uniformly chosen random value \citep{applebaum2006computationally,applebaum2006cryptography}. Decoding of the encoding reveals only the output of the function $f$ but nothing else. 

\citet{applebaum2017garbled} introduced the perfect decomposable and affine randomized encoding (DARE) of some operations in their study. For a randomized encoding to be affine and decomposable, all components of the encoding should be affine functions over the set on which the function is defined and each of these components should depend on only a single input value and a varying number of random values. Here, we give only the encodings that we used, which are addition and multiplication-addition operations.

\begin{definition}[{\bf Perfect RE for Addition \citep{applebaum2017garbled}}] 
\label{def:readd}
Let there be an addition function $t=f(s_1,s_2)=s_1+s_2$ defined over some finite ring $\mathsf{R}$. The following DARE can perfectly encode such a function:
\begin{equation*}
    \hat{f}(s_1,s_2;r) = (s_1+r, s_2-r)
\end{equation*}
where $r$ is a uniformly chosen random value. The decoding can be done by summing up the components of the encoding, and the simulation of the function can be performed by sampling two random values whose sum is $t$.
\end{definition}

\begin{definition}[{\bf Perfect RE for Multiplication-Addition \citep{applebaum2017garbled}}]
\label{def:remuladd}
Let there be a function $t=f(s_1,s_2,s_3)=s_1 \cdot s_2 + s_3$ defined over a ring $\mathsf{R}$. The following DARE function $\hat{t}=\hat{f}(s_1,s_2,s_3;r_1,r_2,r_3,r_4)$ can perfectly encode the function $f$:
\begin{equation*}
    \hat{t} = (s_1 - r_1, r_2 s_1 - r_1 r_2 + r_3, s_2 - r_2, r_1 s_2 + r_4, s_3 - r_3 - r_4)
\end{equation*}
where $r_1,r_2,r_3$ and $r_4$ are uniformly chosen random values. Given the encoding $(c_1,c_2,c_3,c_4,c_5)$, the recovery of $f(s_1,s_2,s_3)$ is done by computing $c_1 \cdot c_3 + c_2 + c_4 + c_5$. In order to simulate $\hat{f}$, one can employ the simulator $\mathsf{Sim}(t;c_1,c_2,c_3,c_4) := (c_1,c_2,c_3,c_4,-c_1c_3+t-c_2-c_4)$.
\end{definition}

In addition to the given DAREs, the authors claim that any arithmetic circuit with logarithmic depth can be encoded by a perfect DARE \cite{applebaum2017garbled}. An example of such an arithmetic circuit computing the dot product of two vectors is given in the Supplement. Taking this into account, we encode the dot product of the vectors by utilizing the aforementioned encodings. Since we only deal with the private computation of the dot product of the vectors, we optimize the generation of the encoding. Let us assume that we have vectors $x,y \in R^D$ where $R$ is a finite ring and $D \in \mathbb{Z}^{+}$. In the dot product computation, we have $D$ multiplication nodes in the circuit and the results of these multiplication nodes are summed up by using the addition nodes. To generate the encoding of the dot product of vectors $x$ and $y$, we first find the largest 2's power which is smaller than $D$, which we represent here as $P$ where $2^q = P$ for $q \in \{\mathbb{Z}^{+} \cup \{0\}\}$ and $P < D \leq 2 \cdot P$. We separate the summation of the first $P$ of those multiplication nodes from the summation of the remaining $D-P$ multiplication nodes by using Definition \ref{def:readd}. We repeat the same procedure for these two parts recursively until we end up with a multiplication node. Once we reach the multiplication node $i$ from an addition node, we utilize the DARE for multiplication-addition given in Definition \ref{def:remuladd} where $s_1=x_i$, $s_2=y_i$ and $s_3$ represents the resulting value of addition/subtraction of the random values separating summations up to that node. The pseudo code of the randomized encoding generation of the dot product of two vectors of size $D$ and the encoding of the sample arithmetic circuit are given in the Supplement.
% Algorithm \ref{algo:re_gen}.

Randomized encoding has two main applications, which are secure computing and parallel cryptography. It is commonly used in multi-party computation (MPC) to minimize the round complexity of MPC protocols \cite{prabhakaran2013secure}. Thus, more efficient MPC protocols can be designed using randomized encoding. 

%Compared to the other methods in the literature such as homomorphic encryption and secure multi-party computation that result in a high overhead due to the use of computationally expensive cryptographic tools, the randomized encoding is faster and more efficient.

%% ---------------------------- Methods --------------------------------------------------
\section{Methods} \label{sec:methods}
In this section, we will first explain the scenario employed in the paper. Then, we introduce the randomized encoding based approach and our proposed framework \fw{}. Later, we will give the security definition as well as the security analysis of it based on the given definition. Finally, we will explain the data that we used.

%%% Old version
%In this section, we will first explain the randomized encoding based approach. Then, we will introduce \fw{}. Both approaches can have multiple input-parties. However, for simplicity, we use a scenario with three input-parties, namely Alice, Bob and Charlie with ids $1$, $2$ and $3$, respectively, and a function-party to describe the approaches. Moreover, we will give the security definition as well as the security analysis of it based on the given definition. Finally, we will explain the data that we used.

\subsection{Scenario} \label{sec:scenario}
We consider a scenario where we have multiple input-parties and a function-party, which computes the dot product of vectors of these input-parties and then trains a kernel-based machine learning algorithm. The real life correspondence of such a scenario would be a study in which a researcher wants to employ the same type of data of different patients collected by multiple hospitals, like cancer subtype discovery. In this scenario, one would like to group cancer patients according to similarities with respect to their omics data. For a new patient the subtype could give first hints about how severe the cancer is and how well the prognosis is regarding potential treatments and life expectancy. Due to patient privacy, such data cannot be shared without any permission process, which can significantly slow down the study. However, a framework, like \fw{}, ensuring the protection of the privacy of patients' data enables the researcher to speed up permission processes and enables studies that would otherwise not be approved due to privacy concerns. While describing the approaches, even though both randomized encoding based approach and \fw{} can have multiple input-parties, for simplicity, we use a scenario with three input-parties, namely Alice, Bob and Charlie with ids $1$, $2$ and $3$, respectively, and a function-party.

\subsection{Randomized Encoding Based Approach} \label{subsec:re_based_app}
To address the aforementioned problem, we first implemented a randomized encoding based approach and applied it to our scenario. In this scenario, each of the input-parties has their own data $X \in R^{f \times n_a}$, $Y \in R^{f \times n_b}$ and $Z \in R^{f \times n_c}$, respectively, where $f$ represents the number of features, $n_x$ represents the number of samples in the corresponding input-party and $R$ is a finite ring. Each pair of input-parties needs to communicate separately, i.e., there will be communication between Alice and Bob, Alice and Charlie, and Bob and Charlie. For simplicity, we will explain only the communication between Alice and Bob. To compute $X^T Y$, they first exchange the size of their own data. Afterwards, Alice generates the scheme of the encoding of the dot product by utilizing the randomized encoding generation algorithm given in the Supplement. By using the resulting encoding scheme, she creates a new set of random values for each possible pair of samples consisting of one sample from Alice and one sample from Bob. This is quite important to protect the relative difference of the features of the input-parties' samples from the function-party. For instance, using the component $s_1 - r_1$ in the encoding, the function-party could learn the relative differences of the input values in the case that the same random value $r_1$ is utilized for more than one pair of samples. Once Alice created all random values, she sends Bob the part of these random values that he will use to encode his own data. Afterwards, both Alice and Bob encode their data by employing the corresponding random values and send the resulting components to the function-party along with the gram matrix of their own samples. To compute the dot product of samples of Alice and Bob, the function-party combines these components according to the decoding described in Definition \ref{def:readd} and \ref{def:remuladd}. Such communication is done between all possible pairs of the input-parties, which means if we have $M$ input-parties, there will be ${M \choose 2}$ communication in total (more detailed communication cost analysis is given in Table \ref{tab:summary_of_approaches}). Once the function-party has all partial gram matrices, it constructs the gram matrix by vertically concatenating the horizontally concatenated partial gram matrices $[X^T X, X^T Y, X^T Z]$, $[Y^T X, Y^T Y, Y^T Z]$ and $[Z^T X, Z^T Y, Z^T Z]$. Then, it can compute the desired kernel matrix, which can be computed via the gram matrix, and train a kernel-based machine learning method. The overview of the dot product computation procedure via the randomized encoding based approach is given in the Supplement. Note that in the supervised scenario the input-parties share the labels of the samples with the function-party in plaintext domain since they do not reveal any extra and sensitive information. However, this could easily be extended if more sensitive labels are supposed to be used in the learning process.

%------------------------------------------------------------------------------------------------

\begin{figure*}[h!tb]
  \centering
  \begin{subfigure}[t]{0.45\linewidth}
  	\captionsetup{width=.49\linewidth}
  	\setlength{\fboxsep}{-0.3pt}%
	\setlength{\fboxrule}{1pt}%
    \fbox{\includegraphics[width=\linewidth]{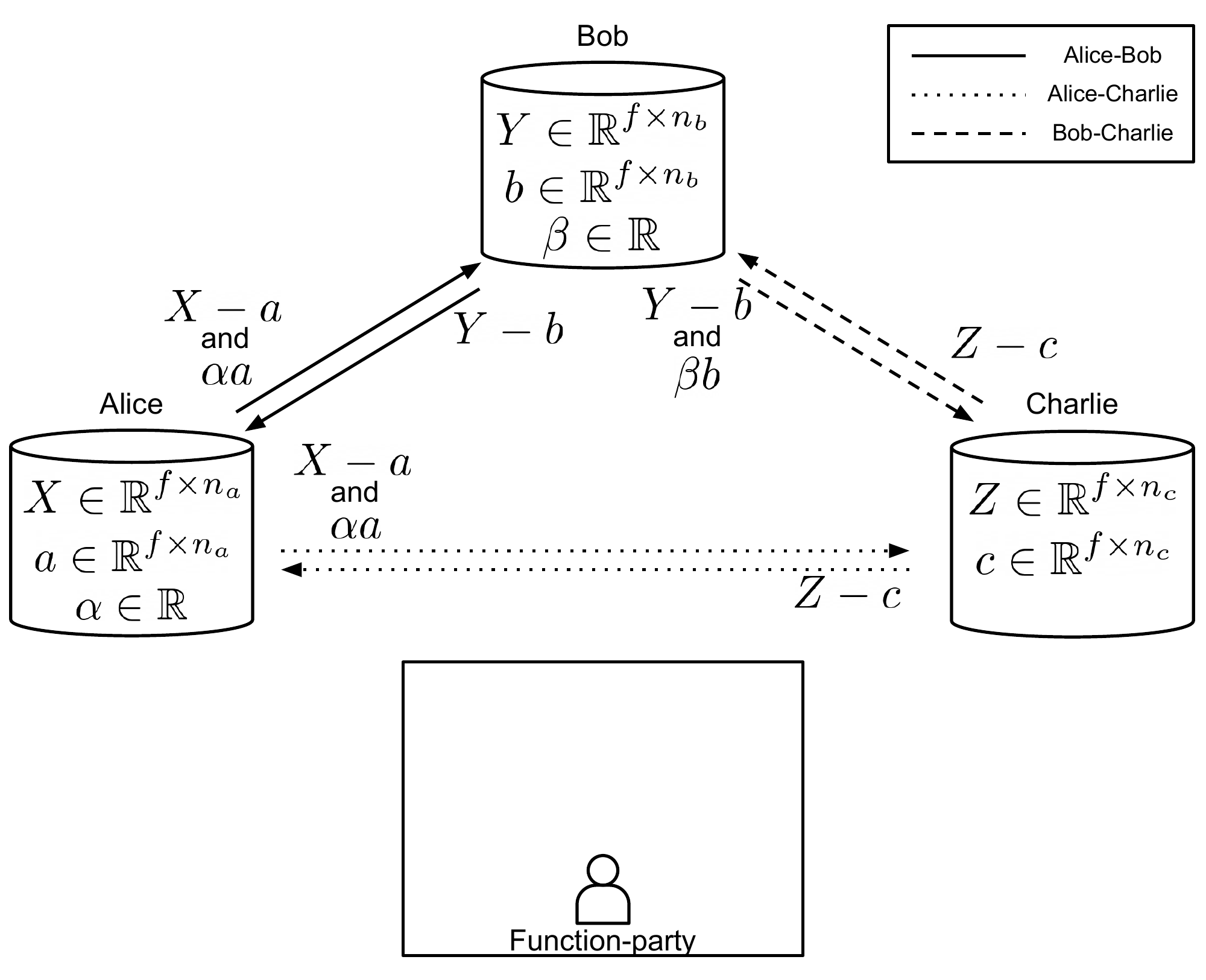}}
    \caption{}
  \end{subfigure}\hspace{15pt}
  \begin{subfigure}[t]{0.45\linewidth}
  	\captionsetup{width=.49\linewidth}
  	\setlength{\fboxsep}{-0.3pt}%
	\setlength{\fboxrule}{1pt}%
    \fbox{\includegraphics[width=\linewidth]{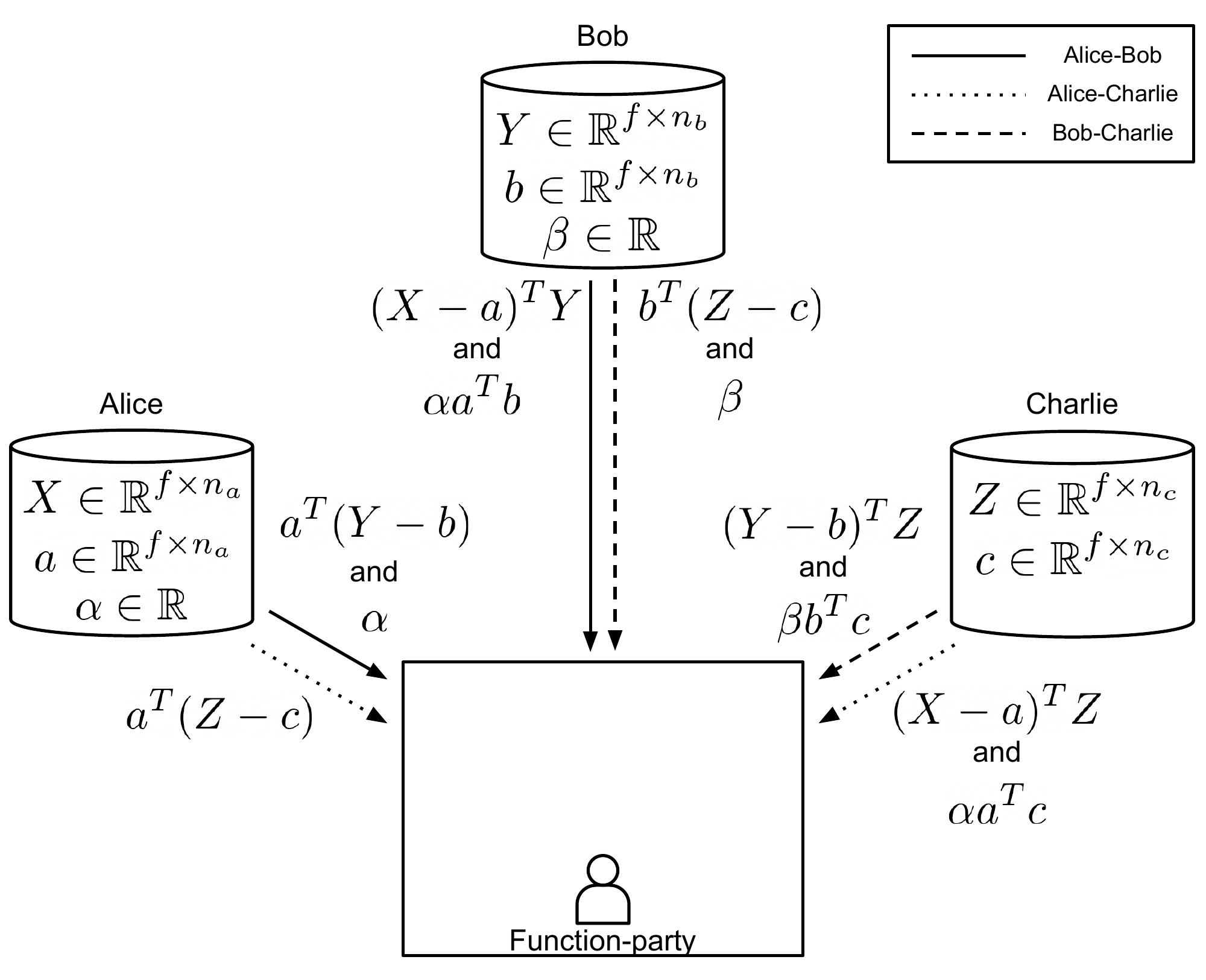}}
    \caption{}
  \end{subfigure} \\
  \begin{subfigure}[t]{0.45\linewidth}
  	\captionsetup{width=.49\linewidth}
  	\setlength{\fboxsep}{-0.3pt}%
	\setlength{\fboxrule}{1pt}%
    \fbox{\includegraphics[width=\linewidth]{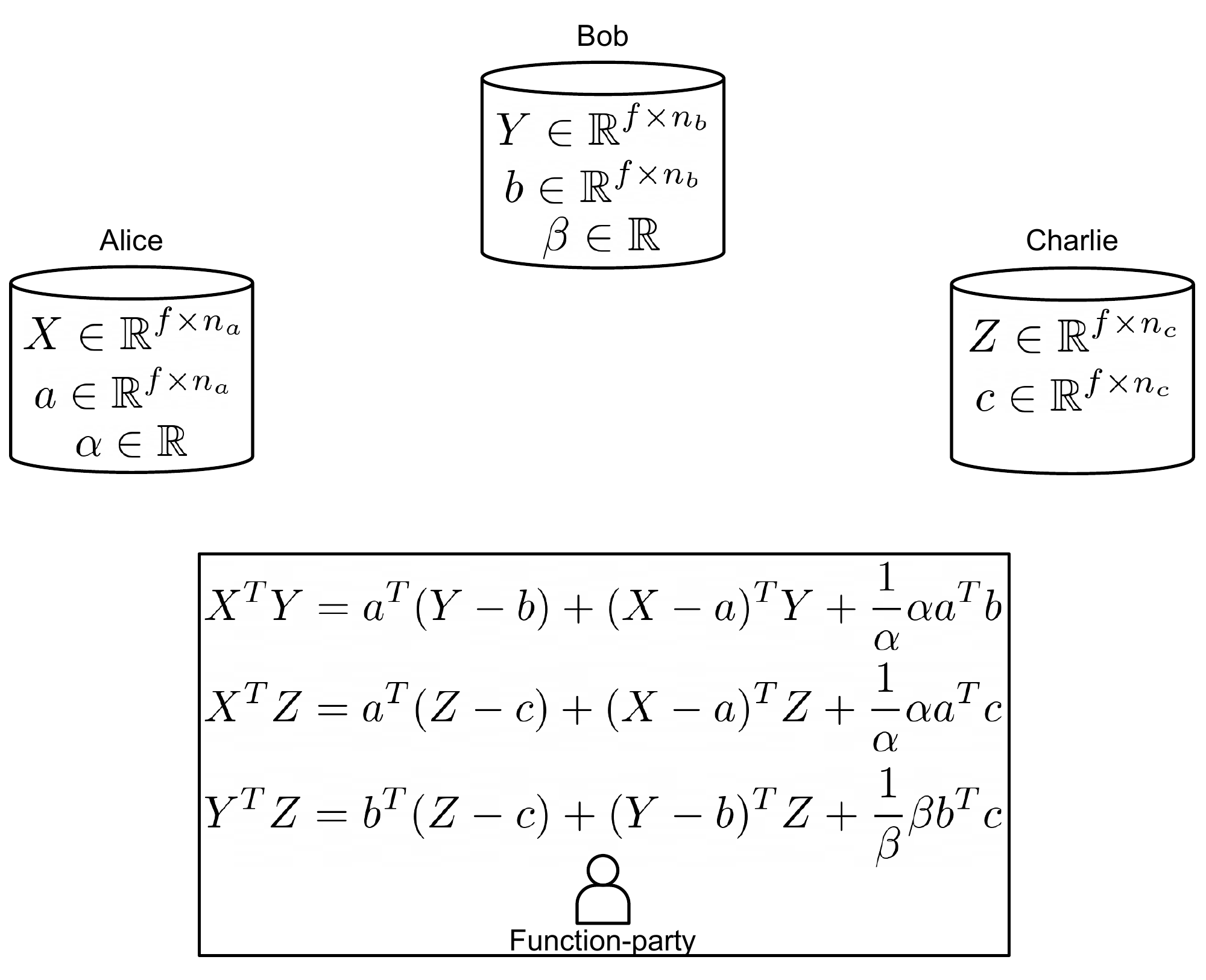}}
    \caption{}
  \end{subfigure}
  \caption{The overview of \fw{} in our scenario is shown. Each dash type corresponds to a specific part of the gram matrix computed by a pair of input-parties. \textbf{(a)} At first, the input-parties exchange their masked input data (e.g. $X - a$) and masked masks (e.g. $\alpha a$) if applicable. \textbf{(b)} Then, they compute the components of all dot products that they are responsible for (e.g. $a^T (Y-b)$) and send them to the function-party along with the mask of the mask (e.g. $\alpha$) if applicable. \textbf{(c)} The function-party computes the dot product based on the corresponding components of the input-parties.}
  \label{fig:esdp_overview}
\end{figure*}

%\subsection{Extended secure dot product (\fw)} \label{subsec:esdp}
% \subsection{Efficient Secure and Private Dot Product}
\subsection{\fw{}}
\label{subsec:esdp}
Due to the high communication cost of the randomized encoding based approach to securely compute the dot product of vectors from multiple input-parties in the function-party, we propose a new efficient and secure framework, which we call \fw{}, based on a new encoding scheme for the dot product computation. In the computation, the input-parties do not learn anything about the data of the other input-parties or the result of any dot product computed by the function-party. Similarly, the function-party learns only the dot product of the data from the input-parties, but nothing else.

For simplicity, we will explain only the computation of the dot product of the data of Alice and Bob in \fw{}. Figure \ref{fig:esdp_overview} depicts the overview of \fw{}. At first, Alice and Bob create matrices of random values $a \in R^{f \times n_a}$ and $b \in R^{f \times n_b}$, respectively, where $R$ is a finite ring. Along with these random valued matrices, Alice also creates a random value $\alpha \in R \setminus \{0\}$. Afterwards, Alice computes $X-a$ and $\alpha a$, and shares them with Bob. In the meantime, Bob computes $Y-b$ and sends it to Alice. Once Alice receives the masked data of Bob, she computes $A_1 = a^{T}(Y-b)$. Meanwhile, Bob computes $B_1 = (X-a)^{T}Y$ and $B_2 = \alpha a^{T} b$. Then, Alice sends $A_1$ and $\alpha$, and Bob sends $B_1$ and $B_2$ to the function-party along with the gram matrix of their own samples, which are $X^T X$ and $Y^T Y$, respectively. At this point, the function-party computes $A_1 + B_1 + \frac{1}{\alpha} B_2$ to obtain the dot product of the data of Alice and Bob, which is $X^T Y$. Such communication is done similarly among all pairs of input-parties. 
In these communications, the input-party with a smaller id becomes {\it ``Alice''} and the other becomes {\it ``Bob''}.
At the end, the function-party has the gram matrix of all samples. Afterwards, the function-party can compute the desired kernel matrix like the RBF kernel matrix, which can be calculated by using Equation \ref{eq:rbfdotproduct}, and train a kernel-based machine learning method utilizing the computed kernel matrix to obtain a prediction model. Note that the input-parties share the labels with the function-party in the plaintext domain because of the same reason that we mentioned earlier.

Table \ref{tab:summary_of_approaches} summarizes the features and the communication cost analysis of \fw{}, the randomized encoding based approach and the approach proposed by \cite{unaletal2019}.

%%%%%%%%%%%%%%%%%%%%%%%%%%%%%%%%%%%%%%%%%%%%%%%%%%%%%%%%%%%%%%%%%%%%%%%%%%%%%%%%%%%%%%%%%%%%%%%%%%%%%%%%%%%%%%%%%

\subsection{Security Definition}
% the thread models and their description
In our proof, we utilize two different adversarial models, which are the {\it semi-honest} adversary model, in other words {\it honest-but-curious}, and the {\it malicious} adversary model. A {\it semi-honest} adversary is a computationally bounded adversary which follows the protocol strictly but also tries to infer any valuable information from the messages seen during the protocol execution. On the other hand, in the {\it malicious} adversary model, a {\it malicious} adversary can arbitrarily deviate from the protocol specification. Although the semi-honest model has more restrictive assumptions than the malicious model, the development of highly efficient privacy preserving protocols is relatively easy under it.

% the setting (num of ips etc.) and assumptions
Let there be $M$ input-parties ($\mathcal{I}_1,...,\mathcal{I}_M$) and a function-party $\mathcal{F}$ in the proposed system. We assume that an adversary is either a semi-honest adversary corrupting a subset of input-parties or a malicious adversary corrupting the function-party. We restrict the collusion between the function-party and the input-parties not to allow the corruption of the function-party and at least one input-party at the same time. Otherwise, an adversary $\mathcal{A}$ who corrupts the function-party and at least one input-party obtains the inputs of all other input-parties. Even though we allow the collusion among input-parties, one might think that it is not so realistic since entities involved, such as medical institutions, lose their reputations if they misbehave in this setting.

% the utilized technique to prove the security of the framework
We use the simulation paradigm \cite{Lindell2017} in our security proofs. In the simulation paradigm, the security is proven by showing that the simulator can simulate the input and the output of a party given the actual input and output such that the simulated input and output cannot be distinguished from the actual ones by an observer. Such an indistinguishability indicates that the parties cannot learn more than what can be learned from their inputs and outputs.

Since the function-party constructs the final output, which is the gram matrix, by using the partial outputs each of which is computed by a pair of input-parties, we can consider these computations as a separate two-party computation. The following notations are used in the security definition:

\begin{itemize}
    \item Let $f = (f_{1}, f_{2})$ be a probabilistic polynomial-time functionality where $f_p$ is the input provided by the $p$-th party to $f$ and let $\pi$ be a two-party protocol for computing $f$.
    \item The view of the $i$-th party $(i \in {1, 2})$ during an execution of $\pi$ over $(x, y)$ is denoted by $v_i^\pi(x, y)$ and equals $(w, r^i,m_1^i,...,m_t^i)$ where $w \in \{x, y\}$, $r^i$ equals the contents of the $i$-th party’s internal random tape and $m^i_j$ represents the $j$-th message that it received.
    \item The output of the $i$-th party during an execution of $\pi$ over $(x, y)$ is denoted by $o^\pi_i(x, y)$ and can be computed from its own view of the execution. We denote the joint output of both parties by $o^\pi(x, y) = (o^\pi_1(x, y),o^\pi_2(x, y))$.
\end{itemize}

\begin{definition}
Let $f = (f_1, f_2)$ be a functionality. We say that a protocol $\pi$ is secure against semi-honest adversaries if there exist probabilistic polynomial time (PPT) simulators $S_1$ and $S_2$ such that: \[(S_1(x, f_1(x, y)), f(x, y)) \overset{c}{\equiv} (v^\pi_1(x, y), o^\pi_1(x, y))\] 
\[(S_2(y, f_2(x, y)), f(x, y)) \overset{c}{\equiv} (v^\pi_2(x, y), o^\pi_2(x, y))\] where $\overset{c}{\equiv}$ denotes the computational indistinguishability. More details can be found in \cite{Oded:2009:FCV:1804390}. 
\end{definition}

\subsection{Security Analysis}
\begin{theorem}
\fw{} is secure against a semi-honest adversary $\mathcal{A}$ who corrupts any subset of input-parties. 
\end{theorem}

\begin{proof}
The proof is provided in the Supplement.
\end{proof}

\begin{theorem}
Assume that the function-party is malicious and does not collude with any input-parties. Then, \fw{} is secure against the malicious function-party $\mathcal{A}$.
\end{theorem}

\begin{proof}
The proof is provided in the Supplement.
\end{proof}

%%%%%%%%%%%%%%%%%%%%%%%%%%%%%%%%%%%%%%%%%%%%%%%%%%%%%%%%%%%%%%%%%%%%%%%%%%%%%%%%%%%%%%%%%%%%%%%%%%%%%%%%%%%%%%%%%

\subsection{Data}
In this section, we briefly explain the datasets that we employed in our supervised and unsupervised learning experiments, respectively.

\textbf{\textit{HIV V3 Loop Sequence Dataset}:}
To predict the personalized treatmet of HIV-infected patients in the supervised learning experiments, we retrieved the HIV V3 loop dataset from \citet{unaletal2019}. It consists of the protein sequence of the viruses as well as their coreceptor usage information. Due to the availability of drugs blocking the human CCR5 coreceptor, which is exclusively used by the most common variant of HIV to enter the cell, identifying the coreceptor usage is very crucial to determine whether or not to use these drugs \cite{Lengauer2007a}. The dataset consists of $642$ samples for the class ``CCR5 only'' and $124$ samples for the class ``OTHER''. The sequence data exists as a one-hot encoded data matrix with $766$ rows and $924$ columns.

\textbf{\textit{Head and Neck Squamous Cell Carcinoma Dataset}:}
We aim to perform the privacy preserving multi-omics dimensionality reduction and clustering on the TCGA data for head and neck squamous cell carcinoma (HNSC) \cite{cancer2015comprehensive} to stratify patients into clinically meaningful subgroups. Therefore, we replicate a recent state-of-the-art study \cite{roder2019web} in a privacy-preserving setting, obtaining the data from the authors. The data consists of $465$ patients with their gene expression (IlluminaHiSeq), DNA methylation (Methylation450k), copy number variation (gistic2), and miRNA expression (IlluminaHiSeq) data. They have $19433$, $57159$, $23817$ and $581$ features, respectively. We also obtained the survival times of the patients.

%% ---------------------------- Results ------------------------------------
\section{Results} \label{sec:results}
%We conducted experiments on the coreceptor prediction problem based on HIV V3 loop sequences and experiments on HNSC cancer patient clustering via \fw{}. 
In order to simulate multiple input-parties, we created a process for each input-party and shared the data among them equally. We also created an additional process to simulate the function-party. All processes communicate with each other over TCP sockets and we assume that the communication is secure. We conducted the experiments on a server which has 512 GB memory, an Intel Xeon E5-2650 processor and a 64-bit operating system. We utilized Python to implement \fw{} and the randomized encoding based approach.

%---------------------------------------------------------------------------------------------------------------
\midsepremove
\begin{table*}[t]
    \centering
    \renewcommand{\arraystretch}{1.7}
    \begin{tabularx}{\textwidth}{CGGFFR}
        \toprule
         & \multicolumn{2}{>{\columncolor{gray!10!white}}c}{Number of IPs} & \multicolumn{3}{>{\columncolor{gray!30!white}}c}{Communication cost} \\
        \hhline{*1{>{\arrayrulecolor{gray!5!white}}-}*5{>{\arrayrulecolor{black}}-}}
        & &  &  &  & \\
        \multirow{-3}{*}{Approach} & \multirow{-2}{*}{\makecell{Two\\IPs}} & \multirow{-2}{*}{\makecell{Three or\\more IPs}} & \multirow{-2}{*}{\makecell{Among\\IPS}} & \multirow{-2}{*}{\makecell{Between\\IPs and FP}} & \multirow{-2}{*}{\makecell{Total}} \\ 
        \midrule
        \pfw{} & Yes & No & $3 R^{f \times n^2}$ \footnotemark[1] & $4 R^{f \times n^2}$ \footnotemark[1] & $7 R^{f \times n^2}$ \footnotemark[1] \\ %\midrule
        RE & Yes & Yes & $4{M \choose 2} R^{f \times n^2}$ & $5 {M \choose 2} R^{f \times n^2}$ & $9 {M \choose 2} R^{f \times n^2}$ \\ %\midrule
        \fw{} & Yes & Yes & $3 {M \choose 2} R^{f \times n}$ & $3 {M \choose 2} R^{n^2}$ & $3 {M \choose 2} ( R^{f \times n} + R^{n^2})$ \\ 
        \bottomrule
    \end{tabularx}
    \caption{The summary of the comparison of the methods utilized in this study from different aspects is given. The ability of handling varying number of input-parties (IP) in the framework proposed by \citet{unaletal2019} (\pfw{}), the randomized encoding based approach (RE) and \fw{} are given in the first part of the table. Moreover, $n$ being the number of samples in each IPs, $M$ being the number of IPs and $f$ being the number of features of samples, where $n,M,f \in \mathbb{Z}^{+}$ and $M \geq 2$, the communication cost analysis of RE and \fw{} in terms of the communication cost among IPs, between IP and the function-party (FP) and the total communication cost are given in the second part of the table. The communication cost analysis of \pfw{}, however, is given without any dependency on $M$ since it can only handle two input-parties scenario. Note that we omit the communication cost of sending the gram matrix of the samples belonging to the same IP since it is fixed for all approaches.}
    \label{tab:summary_of_approaches}
\end{table*}

\subsection{Classification of HIV coreceptor usage}
In these supervised learning experiments, we utilized an SVM with an RBF kernel matrix. We optimized the parameters of the SVM, which are the misclassification penalty $C \in \{2^{-5},2^{-4},\cdots,2^{10}\}$ and the weight $w_1 \in \{2^{0},2^{1},\cdots,2^{5}\}$ of the minority class, and the similarity adjustment parameter $\sigma \in \{2^{-5},2^{-4},\cdots,2^{10}\}$ of the RBF kernel via $5$-fold cross-validation and F1-score. Note that we tuned the parameters outside of the approaches and utilized them in the experiments directly. However, one can employ both \fw{} and the randomized encoding based approach for tuning. To have a fair evaluation, we repeated the optimization step $10$ times with different random folds and conducted separate experiments by utilizing each optimal parameter set. We evaluated the experiments via F1-score and area under receiver operating characteristic curve (AUROC).

\interfootnotelinepenalty=10000
\footnotetext[1]{The communication cost analysis is given after an update on \pfw{} to protect the privacy of relative differences between features of samples. Without any update, the communication costs would become $3 R^{f}$, $4 R^{f \times n}$ and $3 R^{f} + 4 R^{f \times n}$, respectively.}

%%% commented out
\iffalse
    \begin{subfigure}{0.33\linewidth}
        \includegraphics[width=\linewidth]{figs/esdp_both_results.pdf}
        \caption{\textbf{(a)} The result of $10$ repetitions of the experiment by using \fw{} with different optimal parameters, which are determined by utilizing different random folds in cross-validation, is shown. We evaluated the models via F1-score and AUROC. }
        \label{fig:esdp_supervised_results}
    \end{subfigure}
\fi
%%% end of the comment

\begin{figure}[h!tb]
    \centering
    \begin{subfigure}{0.46\linewidth}
        \includegraphics[width=\linewidth]{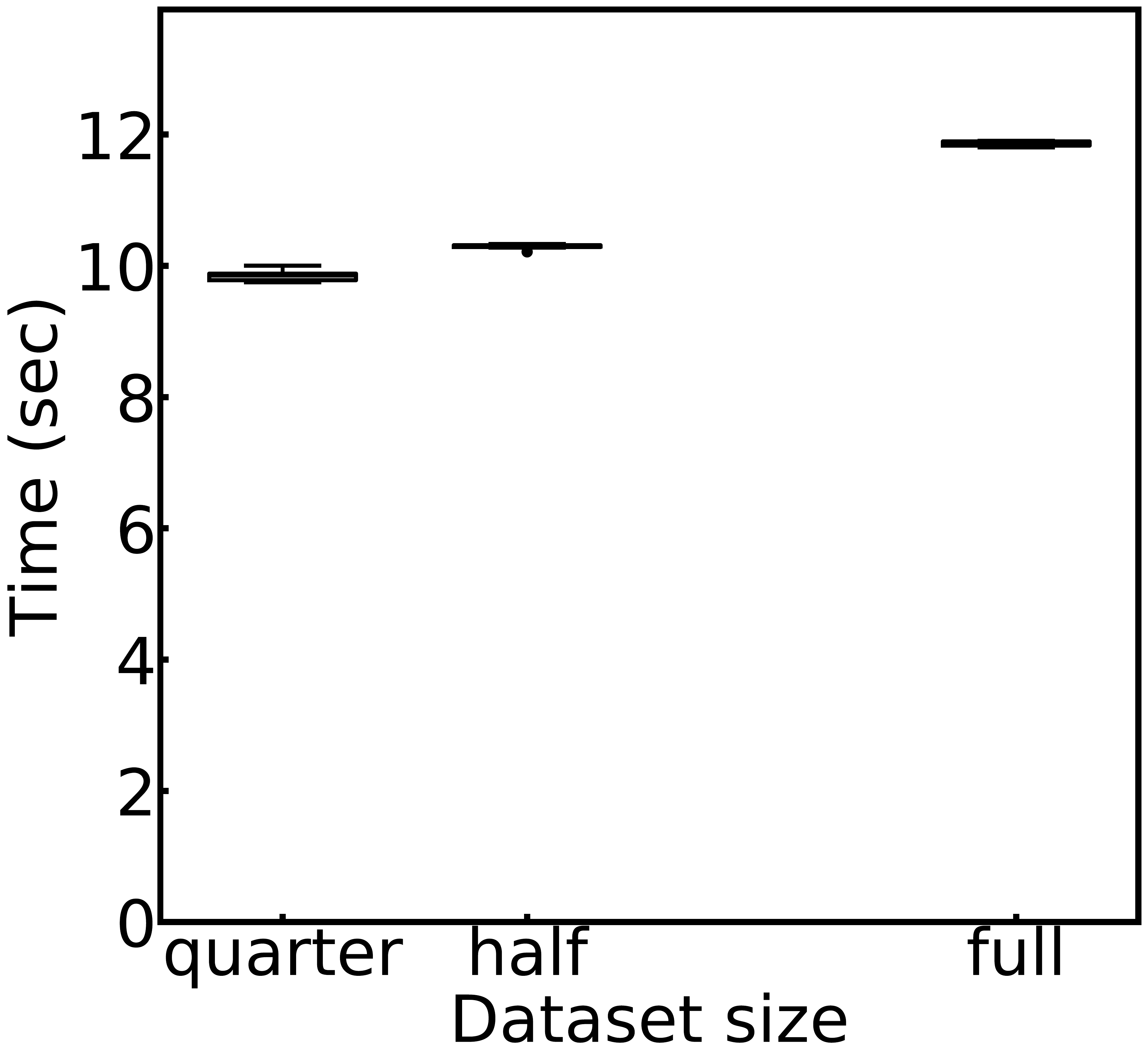}
        \caption{}
        \label{fig:esdp_scalability}
    \end{subfigure}
    \begin{subfigure}{0.485\linewidth}
        \includegraphics[width=\linewidth]{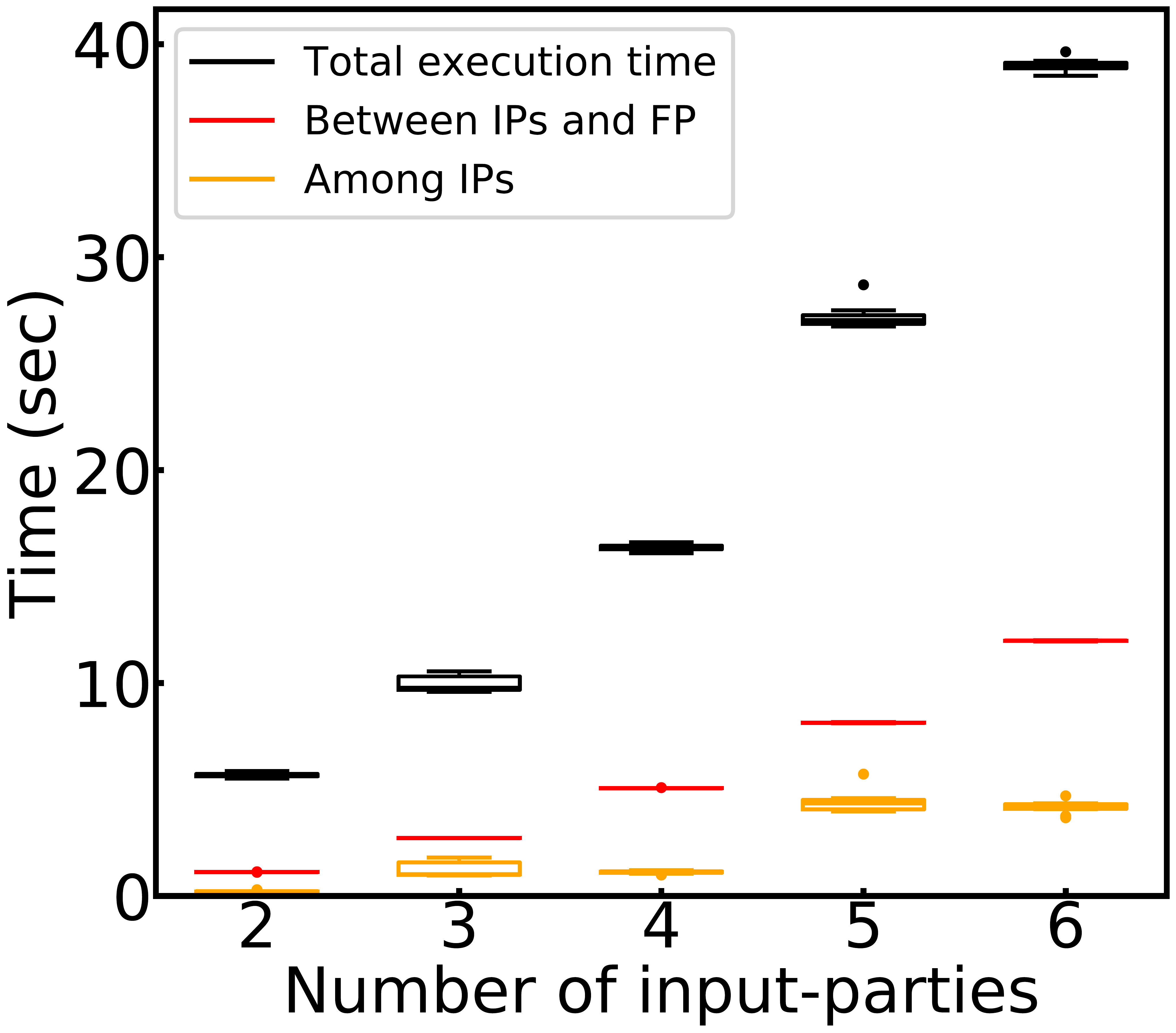}
        \caption{}
        \label{fig:esdp_varying_num_ip}
    \end{subfigure}
    \begin{subfigure}{0.48\linewidth}
        \includegraphics[width=\linewidth]{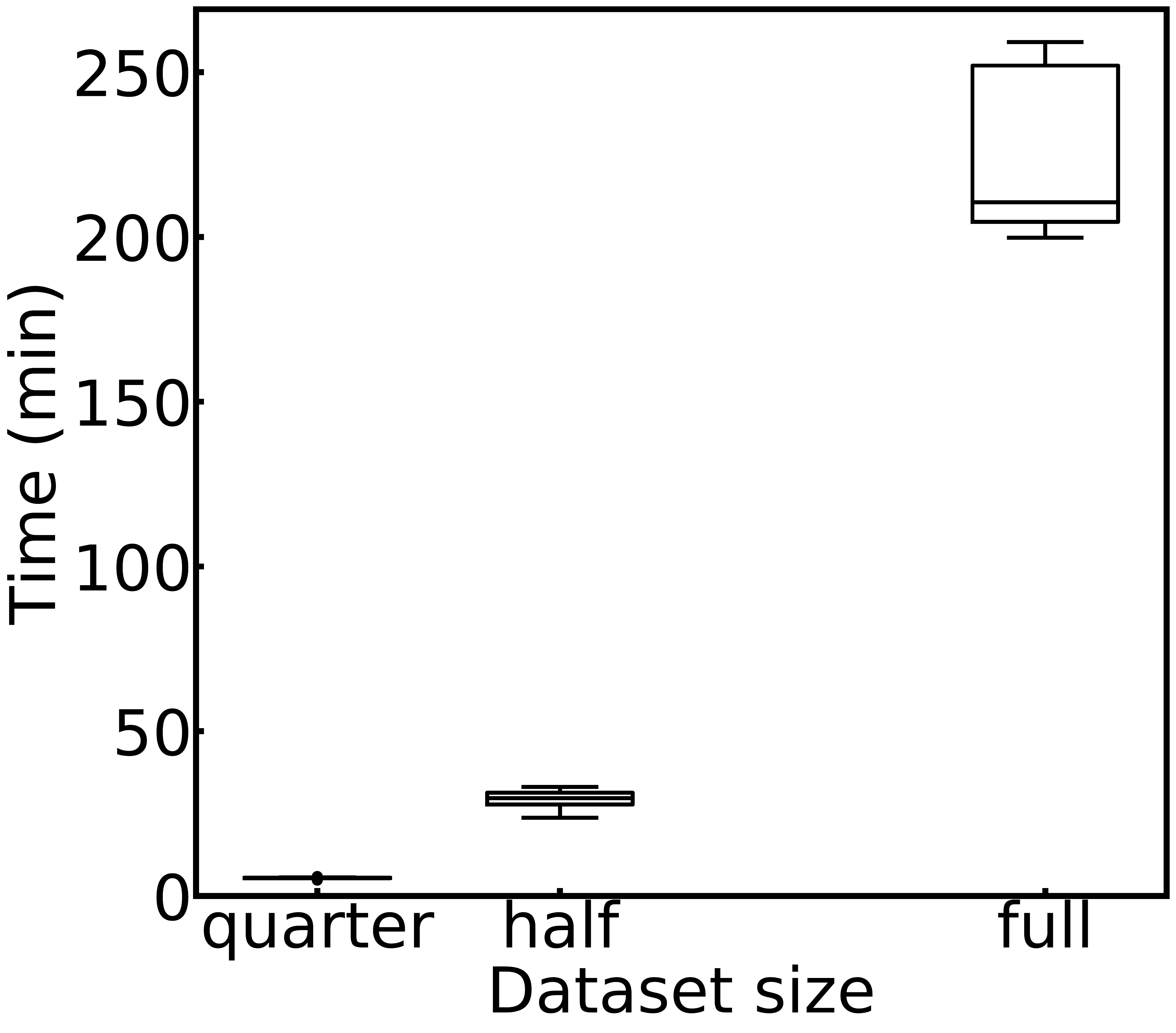}
        \caption{}
        \label{fig:dp_with_re_scalability}
    \end{subfigure}
    \begin{subfigure}{0.48\linewidth}
        \includegraphics[width=\linewidth]{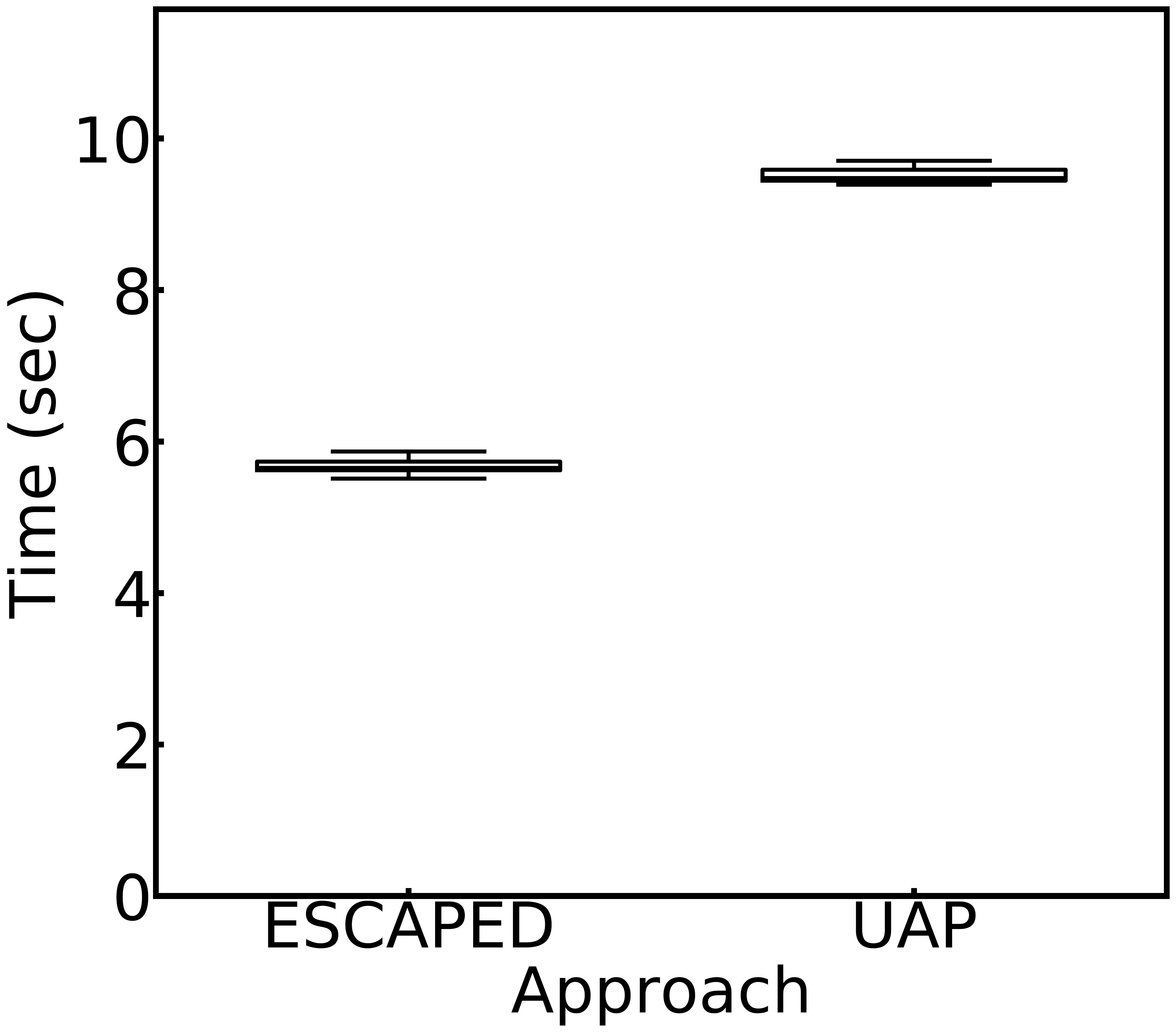}
        \caption{}
        \label{fig:esdp_vs_uap}
    \end{subfigure}
    \caption{ \textbf{(a)} The execution time of \fw{} is shown for varying sizes of the dataset. \textbf{(b)} The analysis of \fw{} for varying number of input-parties in terms of the total execution time, the communication time between the input-parties (IP) and the function-party (FP), and the communication time among IPs is shown. \textbf{(c)} The execution time of the randomized encoding based approach is depicted for different sizes of the dataset. The execution time increases quadratically in parallel to the increment in the dataset size. \textbf{(d)} The execution time comparison of \fw{} and the \pfw{} \cite{unaletal2019} for two input-parties case is given.}
    \label{fig:esdp_result_and_time}
\end{figure}

We utilized our proposed framework, \fw{}, to compute the dot product of samples of three different input-parties on a function-party. Once the function-party has the gram matrix, it computes the RBF kernel matrix based on the optimal $\sigma$ by using Equation \ref{eq:rbfdotproduct}. We separated $20\%$ of the data of each input-party for testing. The function-party trains an SVM model on the rest of the data by employing the optimal parameters $w_1$ and $C$. Then, we tested the model on the test data. At the end, we evaluated the prediction of our model via F1-score and AUROC. We repeated this experiment for each optimal parameter set and we obtained $0.843$ $(\pm 0.013)$ AUROC and $0.615$ $(\pm 0.016)$ F1-score on the average. To demonstrate the scalability of \fw{} in terms of the total dataset size, we conducted experiments in which we used a quarter, a half and the full dataset. The execution time of \fw{} increases almost quadratically with respect to the size of the dataset. Figure \ref{fig:esdp_scalability} shows the trend of the increase in the execution time in parallel to the increment in the dataset size. Furthermore, we analyzed the performance of the framework for varying number of input-parties each of which has the same number of samples. Figure \ref{fig:esdp_varying_num_ip} displays the effect of the number of input-parties involved in the computation on the execution time of various parts. The total execution time and the total communication time between input-parties and the function-party (black and red, respectively) are almost linear. The total communication among input-parties (orange), however, displays a slightly different pattern. Since there is an idle party in each turn of the communication among input-parties when there is an odd number of input-parties, the execution time for the cases having even number of input-parties is almost the same with the case where we have one less input-party.

%%% commented out
\iffalse
\begin{figure}[h!tb]
  \centering
  \begin{subfigure}[t]{0.36\linewidth}
  	\includegraphics[width=\linewidth]{figs/dp_with_re_exe_size_v2.eps}
    \caption{}
    \label{fig:dp_with_re_scalability}
  \end{subfigure}
  \begin{subfigure}[t]{0.36\linewidth}
  	\includegraphics[width=\linewidth]{figs/log_base_comparison.eps}
    \caption{}
    \label{fig:supervised_comparison}
  \end{subfigure}
  \caption{ \textbf{(a)} The execution time of the randomized encoding based approach is depicted for different sizes of the dataset. The execution time increases quadratically in parallel to the increment in the dataset size. \textbf{(b)} The comparison between \fw{} and the randomized encoding (RE) approach in terms of execution time is shown for the full dataset experiments in $log_{10}$-scale.}
  \label{fig:re_figs}
\end{figure}
\fi
%%% end of the comment

%\footnotetext[1]{\label{fn:uap} In case of an update on \pfw{} to protect the privacy of relative differences between features of samples, the communication costs would become $3 R^{f \times n^2}$, $4 R^{f \times n^2}$ and $7 R^{f \times n^2}$, respectively.}

We applied also the randomized encoding based approach to the same scenario. Similar to the experiments with \fw{}, we repeated the whole experiment for each optimal parameter set. Since we obtained exactly the same F1-score and AUROC with \fw{} for the same set of parameters, we demonstrate only the execution time of the randomized encoding based approach for varying size of the dataset in Figure \ref{fig:dp_with_re_scalability}. When we compare the execution time of the randomized encoding based approach to \fw{} for full dataset experiments, the randomized encoding based approach took $1.3 \times 10^4$ $(\pm 1.4 \times 10^{3})$ $sec$ whereas \fw{} took only $1.19 \times 10^{1}$ $(\pm 3.5 \times 10^{-2})$ $sec$. Since the randomized encoding based approach is quite inefficient compared to \fw{}, we did not evaluate it in terms of the number of input-parties. Based on the results and the cost analysis shown in Table \ref{tab:summary_of_approaches}, it is fair to claim that \fw{} is more efficient than the randomized encoding approach and other MPC protocols in the literature.

% here, we may add a paragraph discussing the comparison of E-SDP to our previous framework for only two input-parties scenario.
Even though \citet{unaletal2019} cannot handle three or more input-parties, we compared \fw{} to this framework in case of a scenario with two input-parties. Since we obtained the same results for both methods, we only give the execution time comparison of them. Figure \ref{fig:esdp_vs_uap} shows that \fw{} outperforms their framework. Based on this observation, we can state that \fw{} is more efficient and comprehensive considering its applicability to more than two input-parties as well.

%---------------------------------------------------------------------------------------------------------------

\subsection{Clustering of HNSC cancer patients}
To demonstrate the applicability of \fw{} on unsupervised learning problems on multi-view data, we employed it to determine biologically meaningful patient subgroups of cancers patients. \citet{speicher2015integrating} studied such a problem and suggested a regularized multiple kernel learning algorithm with dimensionality reduction (rMKL-DR). The method was recently evaluated as the best method on a large benchmark study comparing many different methods \cite{rappoport2018multi}. Later, \citet{roder2019web} published the online version of the method called web-rMKL. In that study, one of their use cases is the identification of subgroups of HNSC. To stratify patients into biologically meaningful subgroups, they employed four different data types, which are gene expression, DNA methylation, miRNA expression and copy number variation. They computed one RBF kernel matrix, whose $\gamma$ is chosen based on the rule of thumbs, for each data type and input these kernel matrices to the web-rMKL to obtain the subgroups of patients. They pruned patients whose survival is longer than 5 years. Then, they evaluated the results by survival analysis and obtained a $p=0.0006$ in log-rank test. To show the applicability of \fw{}, we replicated their study in a privacy preserving way. We utilized the same dataset and split it into three input-parties equally. We employed \fw{} to compute the kernel matrix for each data type based on the data belonging to different input-parties. It took $129.17$ $(\pm 3.81)$ $sec$ to compute the required kernel matrices. We, then, input the resulting kernel matrices to web-rMKL with the same parameter choices to cluster patients. We applied the same filters and evaluated the results by survival analysis as they did. At the end, we obtained the same p-value indicating that \fw{} is capable of performing privacy preserving multi-omics dimensionality reduction and clustering. We were unable to conduct these experiments via the randomized encoding based approach due to the excessive memory usage stemming from the inefficiency of the randomized encoding on high dimensional data.

\section{Conclusion} \label{sec:conclusion}
The tension between the unavoidable demand of machine learning algorithms for data and the importance of the privacy of the sensitive information in data urges the researchers to come up with efficient and privacy preserving machine learning algorithms. To address this necessity, we introduced \fw{} to enable the secure and private computation of the dot product in our scenario. In \fw{}, we preserve the privacy of the data in the computation while neither sacrificing the performance of the model nor adding noise. 
We demonstrated the efficiency and applicability of \fw{} on the personalized treatment prediction system of HIV-infected patients and the privacy preserving multi-omics dimensionality reduction and clustering of HNSC patients into biologically meaningful subgroups.
%We demonstrated the efficiency and applicability of \fw{} to a supervised learning problem, that is HIV coreceptor usage prediction, and an unsupervised learning problem, that is stratification of HNSC patients into biologically meaningful subgroups via a privacy preserving multi-omics dimensionality reduction and clustering method.
Also, we implemented and applied the randomized encoding based approach to solve these problems securely. In the supervised learning problem, both approaches yielded the same result in terms of F1-score and AUROC, but \fw{} outperformed the randomized encoding based approach in terms of execution time. In the unsupervised learning case, we replicated the state-of-the-art experiments conducted by \citet{roder2019web} in a privacy preserving way without sacrificing performance. This indicates that \fw{} enables making privacy preserving multi-omics dimensionality reduction and clustering whereas it was not possible to compute the required kernel matrices with the randomized encoding based approach, which is one of the fastest competitors, due to the excessive memory usage. Even though we applied \fw{} to two machine learning methods, it is applicable to any method requiring the dot product of the vectors from multiple sources on a third-party, showing the promise to efficiently make also other learning algorithms privacy preserving. As a future work, other commonly used operations in machine learning algorithms could be included in the framework to extend the scope of it. Furthermore, the interpretability of the resulting model could be improved to allow more sophisticated analyses to be done via the model.

%% ----------------------------- Acknowledgement -------------------------------------
\section*{Acknowledgements}
This study is supported by the DFG Cluster of Excellence “Machine Learning – New Perspectives for Science”, EXC 2064/1, project number 390727645. Furthermore, NP and MA acknowledge funding from the German Federal Ministry of Education and Research (BMBF) within the 'Medical Informatics Initiative' (DIFUTURE, reference number 01ZZ1804D).\vspace*{-12pt}

\section*{Ethics Statement}
Thanks to the promising results of \fw{}, our study could open up new collaboration opportunities among hospitals, universities, institutes, data centers and many other entities with faster permission processes by providing secure and private computation of dot product enabling not only the kernel-based learning algorithms but also other methods requiring the dot product. This would help to speed up healthcare research helping humanity and the world in general. We could not think of a negative ethical impact of our work.

\bibliography{esdp}

\end{document}

\maketitle

\section{Randomized Encoding Generation Algorithm}
The pseudo code of the randomized encoding generation of the dot product of two vectors of size $D$ is given in Algorithm \ref{algo:re_gen}.
\begin{algorithm}[h!tb]
\caption{Randomized Encoding Generation for Dot Product}\label{algo:re_gen}
\hspace*{\algorithmicindent} \textbf{Input} \\ \hspace*{\algorithmicindent} \textit{eX}: the list of the indices of the random values for X \\
\hspace*{\algorithmicindent} \textit{eY}: the list of the indices of the random values for Y \\
\hspace*{\algorithmicindent} \textit{eO}: the list of the indices of the random values for the offline part \\
\hspace*{\algorithmicindent} \textit{eOS}: the list of the sign of the random values of the offline part \\
\hspace*{\algorithmicindent} \textit{R}: the starting value of the random values that will be generated \\
\hspace*{\algorithmicindent} \textbf{Output} \\
\hspace*{\algorithmicindent} \textit{R}: the latest value of the generated random values
\begin{algorithmic}[1]
\Procedure{REGen}{eX, eY, eO, eOS, R}
\State $\textit{D} \gets \text{length of }\textit{eX}$
\If{$\textit{D} = 1$}
\State $\textit{eX}[0] \gets [0, \textit{R+1}, \textit{R}, \textit{R}, \textit{R+1}, \textit{R+2}]$  \Comment{In the generated random values afterwards, the first one, indicated by 0 here, is always 1}
\State $\textit{eY}[0] \gets [0, \textit{R}, \textit{R+1}, \textit{R+3}]$
\State $\textit{eO}[0] \gets [\textit{eO}[0], \textit{R+2}, \textit{R+3}]$
\State $\textit{eOS}[0] \gets [\textit{eOS}[0], -1, -1]$
\State $\textit{R} \gets \textit{R} + 4$
\State \Return $\textit{R}$
\EndIf
\State $q \gets \text{argmax}_q \, (2^q < \textit{D})$ \Comment{$q \in \{\mathbb{Z}^+ \cup \{0\}\}$}
\State $P \gets 2^q$

\State $\textit{eO}[0] \gets [\textit{eO}[0], \textit{R}]$
\State $\textit{eOS}[0] \gets [\textit{eOS}[0], +1]$ \Comment{+1 represents positive sign}

\State $\textit{eO}[P] \gets [\textit{eO}[P], \textit{R}]$
\State $\textit{eOS}[P] \gets [\textit{eOS}[P], -1]$ \Comment{-1 represents negative sign}

\State $\textit{R} \gets \textit{R} + 1$

\State $\textit{R} \gets REGen(\textit{eX}[0:\textit{P}],\textit{eY}[0:\textit{P}],\textit{eO}[0:\textit{P}], \textit{eOS}[0:\textit{P}],\textit{R})$

\State $\textit{R} \gets REGen(\textit{eX}[\textit{P}:D], \textit{eY}[\textit{P}:D], \textit{eO}[\textit{P}:D], \textit{eOS}[\textit{P}:D],\textit{R})$
\State \Return \textit{R}
\EndProcedure
\end{algorithmic}
\end{algorithm}

\newpage

\section{Sample Encoding}
\begin{figure}[h!tb]
	\centering
	\includegraphics[width=0.5\linewidth]{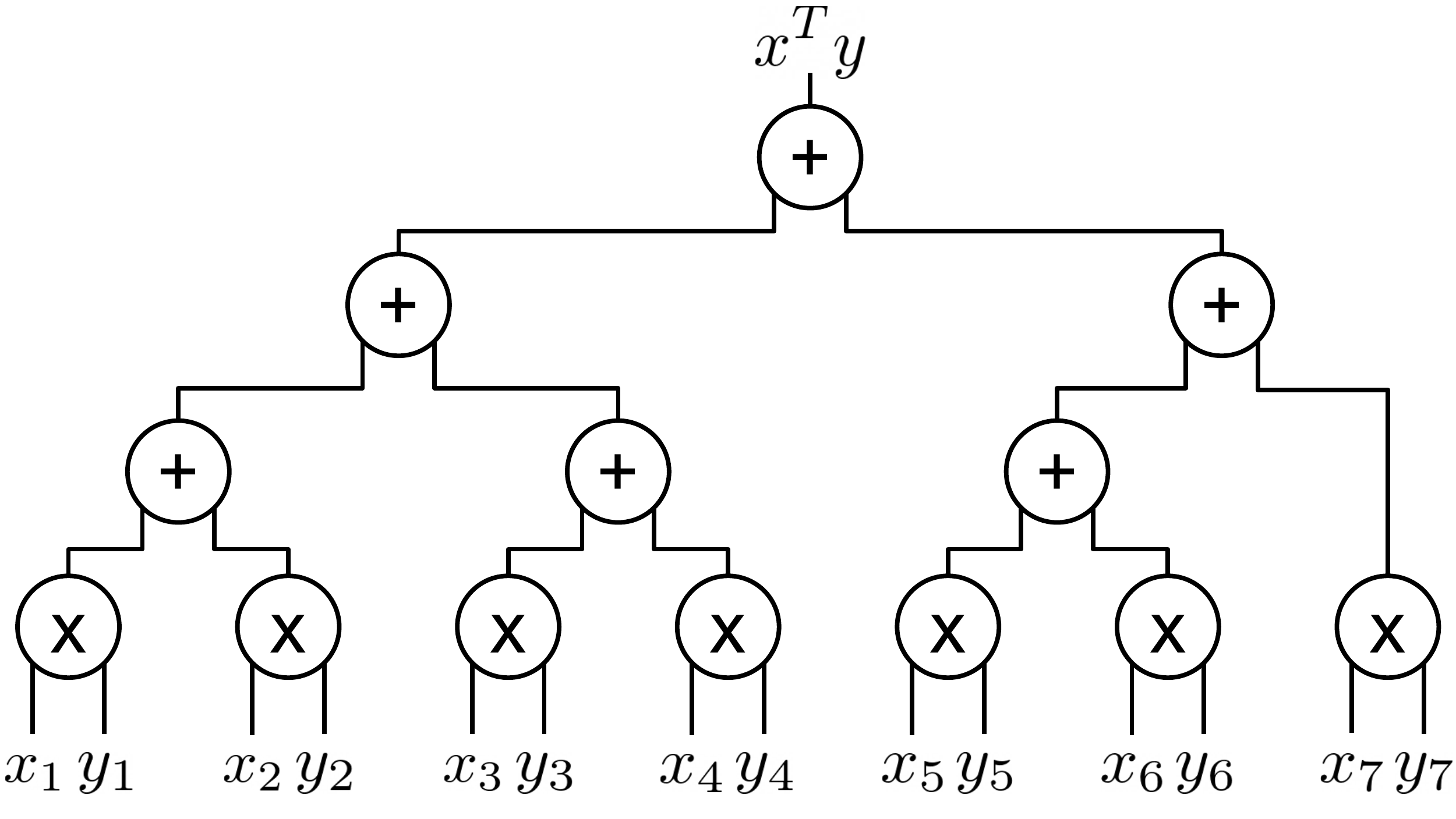}
	\caption{An arithmetic circuit to compute the dot product of the vectors $x$ and $y$ where $x,y \in \mathbb{R}^7$.}
	\label{fig:sample_circuit}
\end{figure}

A sample arithmetic circuit computing the dot product of two vectors is given in Figure \ref{fig:sample_circuit}, where we use two vectors of size $7$. The encoding of the corresponding circuit is given in Equation \ref{eq:sample_encoding}. 

\begin{equation} \label{eq:sample_encoding}
    \begin{split}
        \hat{f}(x,y,R) =  \bigg(& x_1 \begin{bmatrix} 1 \\ r_7 \end{bmatrix} + \begin{bmatrix} -r_6 \\ -r_6 r_7 + r_8 \end{bmatrix},  y_1 \begin{bmatrix} 1 \\ r_6 \end{bmatrix} + \begin{bmatrix} -r_7 \\ r_9 \end{bmatrix}, r_0 + r_1 + r_3 - r_8 -r_9, \\ 
        & x_2 \begin{bmatrix} 1 \\ r_{11} \end{bmatrix} + \begin{bmatrix} -r_{10} \\ -r_{10} r_{11} + r_{12} \end{bmatrix}, y_2 \begin{bmatrix} 1 \\ r_ {10} \end{bmatrix} + \begin{bmatrix} -r_{11} \\ r_{13} \end{bmatrix}, -r_3 - r_{12} + r_{13}, \\ 
        & x_3 \begin{bmatrix} 1 \\ r_{15} \end{bmatrix} + \begin{bmatrix} -r_{14} \\ -r_{14} r_{15} + r_{16} \end{bmatrix}, y_3 \begin{bmatrix} 1 \\ r_ {14} \end{bmatrix} + \begin{bmatrix} -r_{15} \\ r_{17} \end{bmatrix}, -r_1 + r_4 - r_{16} + r_{17}, \\ 
        & x_4 \begin{bmatrix} 1 \\ r_{19} \end{bmatrix} + \begin{bmatrix} -r_{18} \\ -r_{18} r_{19} + r_{20} \end{bmatrix}, y_4 \begin{bmatrix} 1 \\ r_ {18} \end{bmatrix} + \begin{bmatrix} -r_{19} \\ r_{21} \end{bmatrix}, -r_4 - r_{20} - r_{21}, \\ 
        & x_5 \begin{bmatrix} 1 \\ r_{23} \end{bmatrix} + \begin{bmatrix} -r_{22} \\ -r_{22} r_{23} + r_{24} \end{bmatrix}, y_5 \begin{bmatrix} 1 \\ r_ {22} \end{bmatrix} + \begin{bmatrix} -r_{23} \\ r_{25} \end{bmatrix}, -r_0 + r_2 + r_5 - r_{24} - r_{25}, \\ 
        & x_6 \begin{bmatrix} 1 \\ r_{27} \end{bmatrix} + \begin{bmatrix} -r_{26} \\ -r_{26} r_{27} + r_{28} \end{bmatrix}, y_6 \begin{bmatrix} 1 \\ r_ {26} \end{bmatrix} + \begin{bmatrix} -r_{27} \\ r_{29} \end{bmatrix}, -r_5 - r_{28} - r_{29}, \\ 
        & x_7 \begin{bmatrix} 1 \\ r_{31} \end{bmatrix} + \begin{bmatrix} -r_{30} \\ -r_{30} r_{31} + r_{32} \end{bmatrix}, y_7 \begin{bmatrix} 1 \\ r_ {30} \end{bmatrix} + \begin{bmatrix} -r_{31} \\ r_{33} \end{bmatrix}, -r_2 - r_{32} - r_{33} \bigg)
    \end{split}
\end{equation}

Let $\hat{f}(x,y,R)$ be $\bigg( \begin{bmatrix} c_1 \\ c_2 \end{bmatrix}, \begin{bmatrix} c_3 \\ c_4 \end{bmatrix}, c_5, \dotsc, \begin{bmatrix} c_{31} \\ c_{32} \end{bmatrix}, \begin{bmatrix} c_{33} \\ c_{34} \end{bmatrix}, c_{35} \bigg)$, decoding of $\hat{f}$ is done by computing the following:
\begin{equation*}
    f(x,y) = x^T y = \sum_{i \in S} c_i * c_{i+2} + c_{i+1} + c_{i+3} + c_{i+4}
\end{equation*}
where $S = \{1,6,11,16,21,26,31\}$ is the set of indices indicating the beginning of each encoded multiplication $x_i \cdot y_i$ for $i \in \{1, 2, \dotsc, 7\}$.

\newpage

\section{The Scheme of Randomized Encoding Based Approach}
\begin{figure*}[h!tb]
  \centering
  \begin{subfigure}[t]{0.45\linewidth}
  	\captionsetup{width=.9\linewidth}
  	\setlength{\fboxsep}{-0.3pt}%
	\setlength{\fboxrule}{1pt}%
    \fbox{\includegraphics[width=\linewidth]{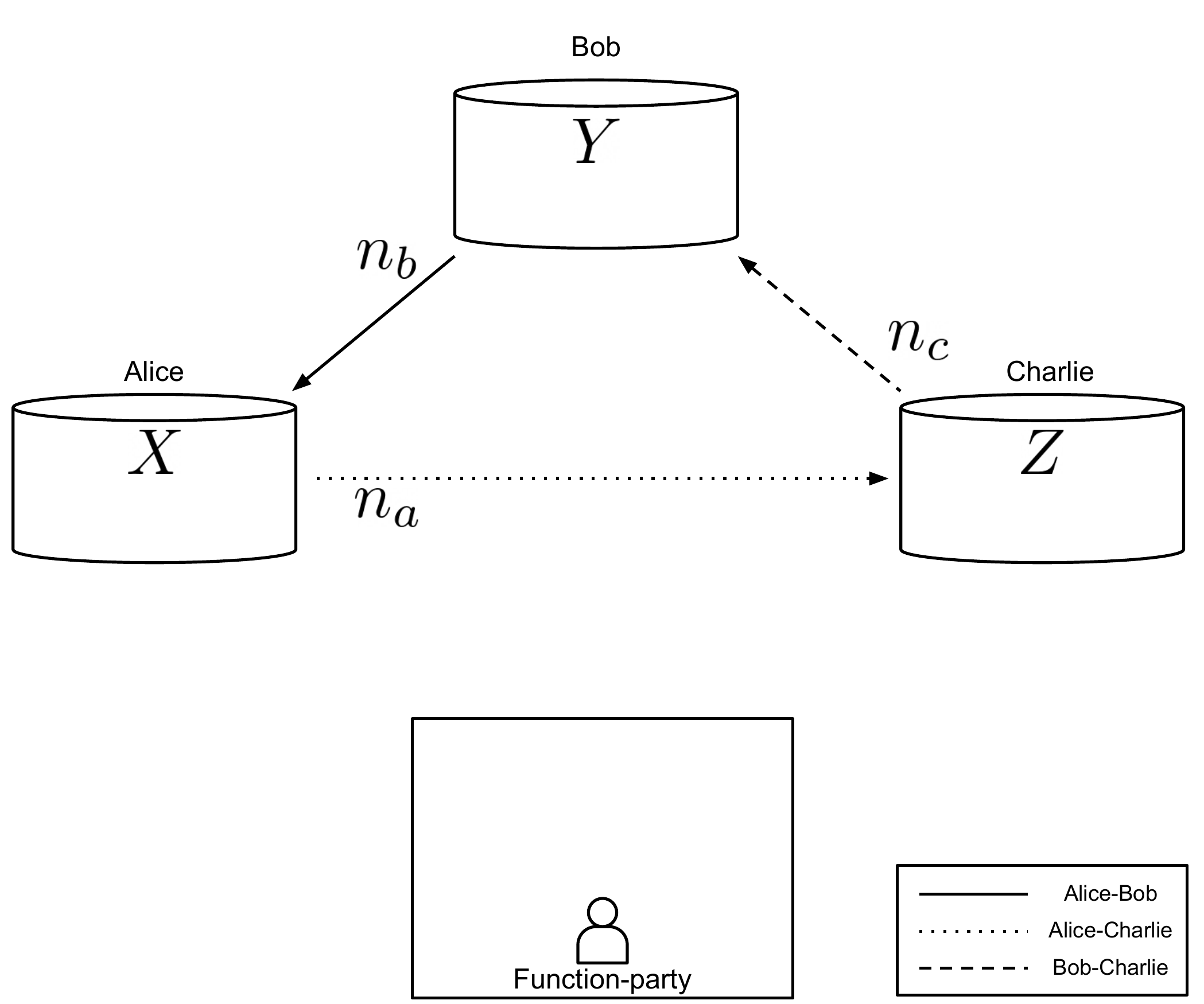}}
     \caption{}
  \end{subfigure}\hspace{15pt}
  \begin{subfigure}[t]{0.45\linewidth}
  	\captionsetup{width=.9\linewidth}
  	\setlength{\fboxsep}{-0.3pt}%
	\setlength{\fboxrule}{1pt}%
    \fbox{\includegraphics[width=\linewidth]{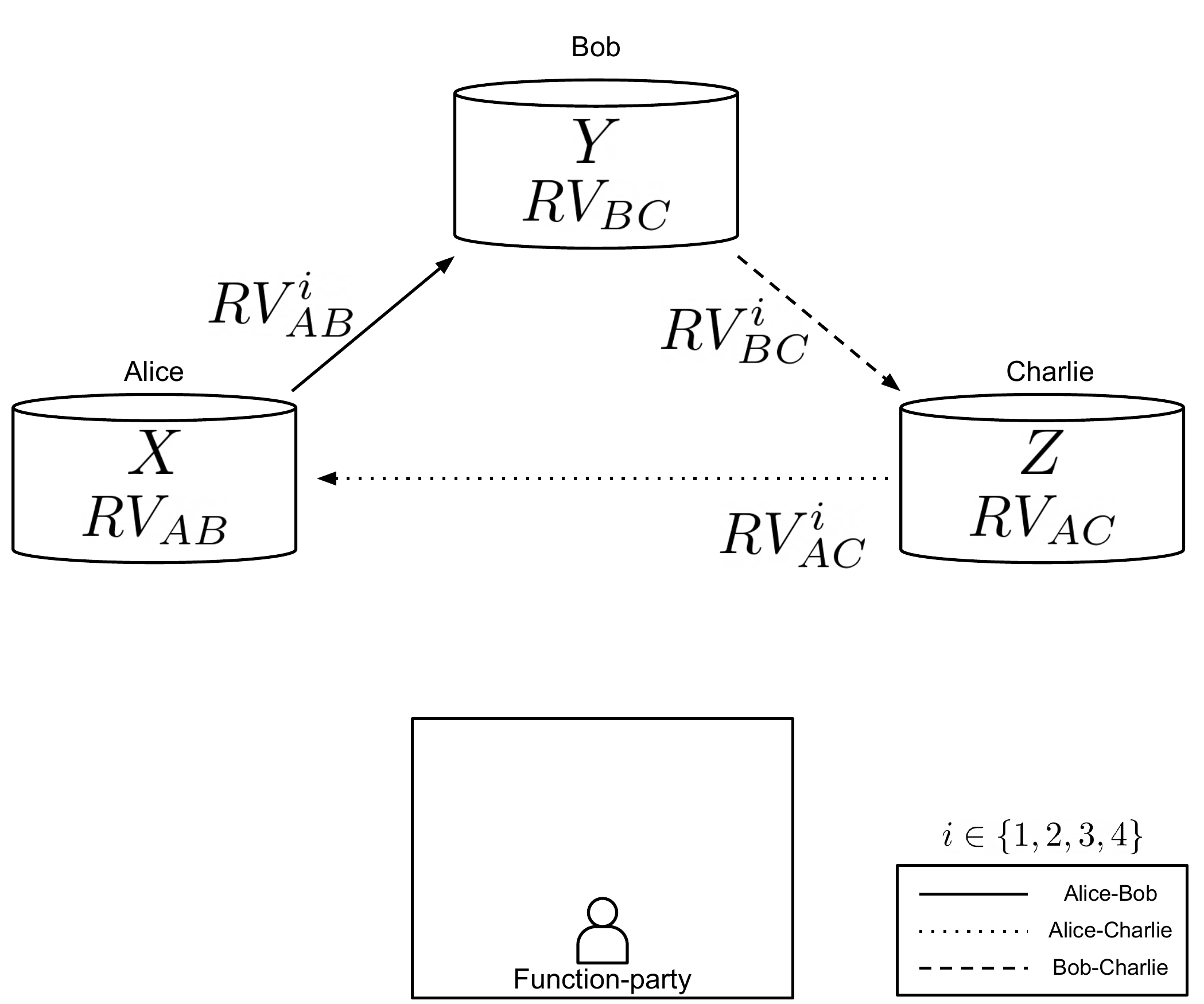}}
    \caption{}
  \end{subfigure}\\
  \begin{subfigure}[t]{0.45\linewidth}
  	\captionsetup{width=.9\linewidth}
  	\setlength{\fboxsep}{-0.3pt}%
	\setlength{\fboxrule}{1pt}%
    \fbox{\includegraphics[width=\linewidth]{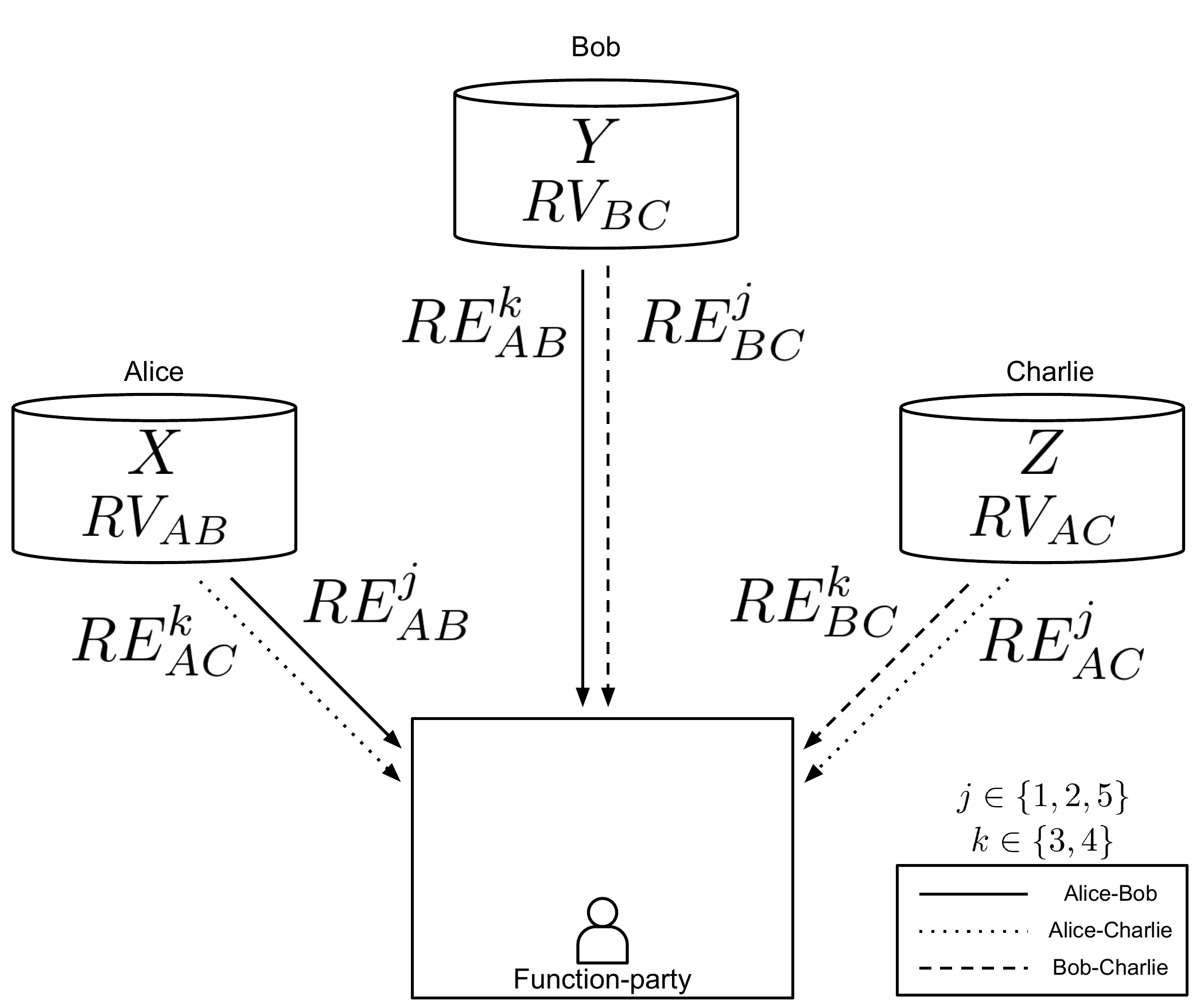}}
    \caption{}
  \end{subfigure}\hspace{15pt}
  \begin{subfigure}[t]{0.45\linewidth}
  	\captionsetup{width=.9\linewidth}
  	\setlength{\fboxsep}{-0.3pt}%
	\setlength{\fboxrule}{1pt}%
    \fbox{\includegraphics[width=\linewidth]{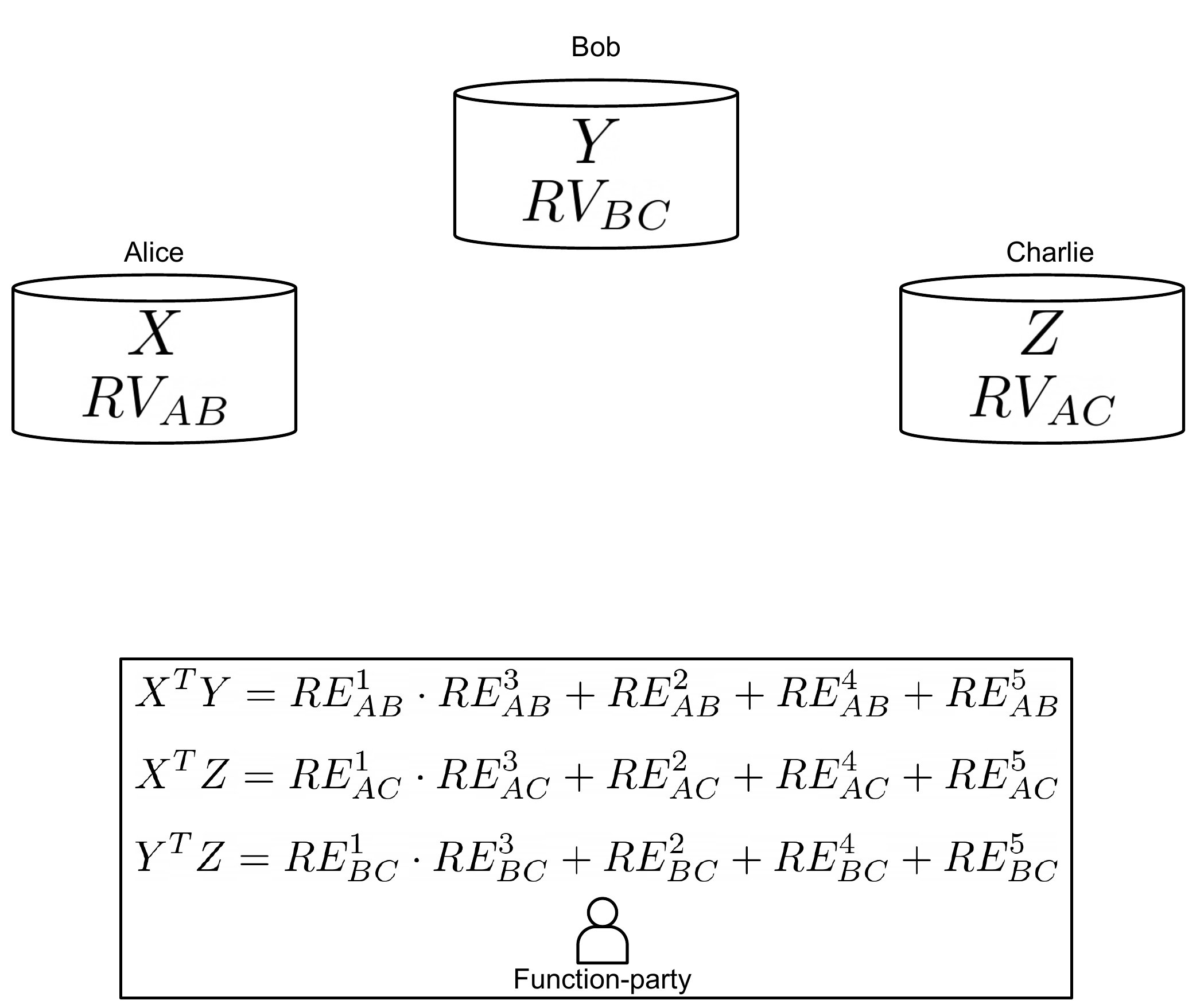}}
    \caption{}
  \end{subfigure}\\
  \begin{subfigure}[t]{0.8\linewidth}
  	\captionsetup{width=.9\linewidth}
  	\setlength{\fboxsep}{0pt}%
	\setlength{\fboxrule}{1pt}%
    \fbox{\includegraphics[width=\linewidth]{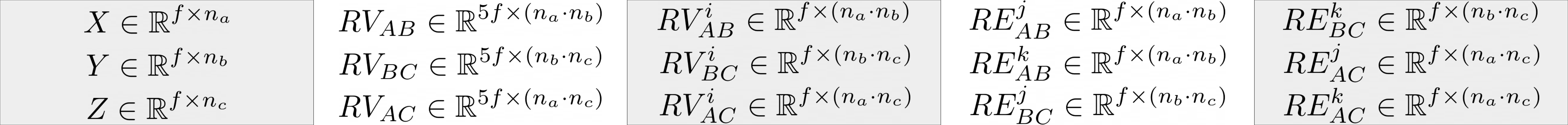}}
    \caption{}
  \end{subfigure}
  \caption{The overview of the randomized encoding based approach to compute the dot product of samples from multiple input-parties is depicted. Each dash type corresponds to a specific part of the gram matrix computed by a pair of input-parties. \textbf{(a)} At first, the input-parties exchange their number of samples. \textbf{(b)} Then, they generate all the random values and send the required ones to the corresponding input-party. \textbf{(c)} Afterwards, they compute their components in the encoding and share them with the function-party. \textbf{(d)} Finally, the function-party computes the dot product of the samples. \textbf{(e)} The dimension of the matrices are shown separately for better readability.}
  \label{fig:re_approach}
\end{figure*}

The scheme of the randomized encoding based approach to compute the gram matrix is depicted in Figure \ref{fig:re_approach}.

\newpage

\section{Security Proof}
\begin{theorem}
Efficient secure and private dot product framework (\fw{}) is secure against a semi-honest adversary $\mathcal{A}$ who corrupts any subset of input-parties. 
\end{theorem}

\begin{proof}
According to the definition of the semi-honest adversary, $\mathcal{A}$ cannot deviate from the protocol description. Thus, $\mathcal{A}$ has to use the real inputs of the corrupted input-parties. Based on this information, it is easy to prove the correctness of \fw{}. For simplicity, we present our proof using a scenario with three input-parties ($\mathcal{I}_1,\mathcal{I}_2,\mathcal{I}_3$). The private data of $\mathcal{I}_1,\mathcal{I}_2$, and $\mathcal{I}_3$ are $X$, $Y$, and $Z$, respectively, and the function-party $\mathcal{F}$ wants to learn $X^T Y, X^T Z$, and $Y^T Z$. Equation \ref{eq:correctness} shows the correctness of \fw{}. 

\begin{equation} \label{eq:correctness}
\begin{split}
X^TY & = a^T(Y-b)+(X-a)^TY + \frac{1}{\alpha}\alpha a^Tb \\
 & = a^TY-a^Tb+X^TY-a^TY+a^Tb \\
 & = X^TY\\
 X^TZ & = a^T(Z-c)+(X-a)^TZ + \frac{1}{\alpha}\alpha a^Tc \\
 & = a^TZ-a^Tc+X^TZ-a^TZ+a^Tc \\
 & = X^TZ\\
 Y^TZ & = b^T(Z-c)+(Y-b)^TZ + \frac{1}{\beta}\beta b^Tc \\
 & = b^TZ-b^Tc+Y^TZ-b^TZ+b^Tc \\
 & = Y^TZ
\end{split}
\end{equation}

After showing the correctness, we now prove the security of \fw{}. $\mathcal{I}_1,\mathcal{I}_2$, and $\mathcal{I}_3$ learn ($Y-b$, $Z-c$), ($X-a$, $\alpha a$, $Z-c$), and ($X-a$, $\alpha a$, $Y-b$, $\beta b$), respectively. These values are generated by using random values $\alpha$ and $\beta$, and matrices of uniformly chosen random values, $a,b$ and $c$. Thus, they can be perfectly simulated by matrices of uniformly random values, which indicates the security of \fw{}. More clearly, an input-party gets the randomly masked input data of other input-parties as well as their randomly masked masks. For instance, $\mathcal{I}_2$ receives $(X-a), (Z-c)$ and $\alpha a$. Since both the mask of the mask, $\alpha$, and the mask of the data, $a$ and $b$, are uniformly random, it is impossible for $\mathcal{I}_2$ to learn the input data of other input-parties, which are $X$ and $Z$ in this case.
\end{proof}

\begin{theorem}
Assume that the function-party is malicious and does not collude with any input-parties. Then, \fw{} is secure against the malicious function-party $\mathcal{A}$.
\end{theorem}

\begin{figure}[H]
    \centering
    \includegraphics[width=0.4\linewidth]{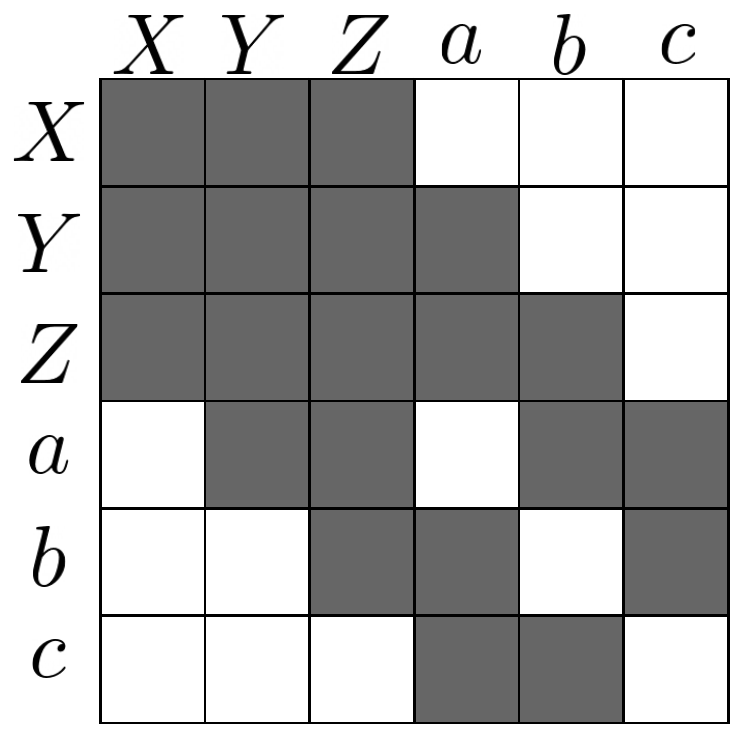}
    \caption{The computed incomplete gram matrix in which only the shaded parts are available.}
    \label{fig:incomplete_gram_matrix}
\end{figure}

\begin{proof}
Since the function-party does not have any input, it cannot change the output of any input-parties. This guarantees and proves the correctness of \fw{} against the malicious function-party. 

The security of the framework from the perspective of the function-party leans on two facts. The first one is that the number of features of the input data is unknown for the function-party. $\mathcal{A}$ cannot be sure which vector space the samples are coming from. This makes unique prediction impossible. The second one is that $\mathcal{A}$ gets the gram matrix of input data among input-parties, that is $X^T Y, X^T Z$ and $Y^T Z$, the gram matrix of their own samples, that is $X^T X, Y^T Y$ and $Z^T Z$, the gram matrix of the input data and the random mask, which are $a^T Y, a^T Z$ and $b^T Y$, and the gram matrix of some of the random masks, more precisely $a^T b, a^T c$ and $b^T c$. They form an incomplete gram matrix $\widetilde{K} = D^T D$, where $D = [X,Y,Z,a,b,c]$ (see Figure \ref{fig:incomplete_gram_matrix}). Even if the number of features is known by $\mathcal{A}$ and the complete version $K = D^T D$ is available, one can come up with multiple matrices satisfying the gram matrix K. Assume that there is a rotation matrix $R \in \mathbb{R}^{N \times N}$ where $N = 2(n_a + n_b + n_c)$. Then, we can compute a matrix $E$ such that $E = R^{-1}D$. Hence, we can express $D$ as $D=RE$. Then, the computation of $K = D^T D$ becomes as follows:
\begin{equation} \label{eq:non-uniqueness-proof}
\begin{split}
K & = D^T D \\
 & = (RE)^T (RE) \\
 & = E^T R^T R E \\
 & = E^T R^{-1} R E \\
 & = E^T E
\end{split}
\end{equation}
due to the property of the rotation matrices, which is $R^{-1}=R^T$. Based on this observation, we can say that for every new rotation matrix $\Theta \in \mathbb{R}^{N \times N}$ there exists a new matrix $\beta = \Theta^{-1} D$ satisfying $K = \beta^T \beta$. Since \citet{aguilera2004general} demonstrated a way to generate rotation matrices for any dimension, one cannot obtain a unique matrix satisfying $K$. To be exact, one cannot recover $D$ from $K=D^T D$, which implies that it is not possible to recover $D$ from $\widetilde{K} = D^T D$ either.
\end{proof}

\bibliography{esdp}